\documentclass[sigconf]{acmart}

\AtBeginDocument{%
  }

\usepackage{soul}
\usepackage{url}
\usepackage[utf8]{inputenc}
\usepackage{caption}
\usepackage{amsthm}
\usepackage{bbm}
\usepackage{booktabs}
\usepackage{algorithm}
\usepackage{algpseudocode}

\usepackage[switch]{lineno}
\usepackage{xcolor}
\usepackage{amsmath}
\usepackage{multirow}
\setcopyright{acmlicensed}
\copyrightyear{2025}
\acmYear{2025}
\acmDOI{10.1145/3746252.3761009}
\usepackage{tikz}
\DeclareMathOperator*{\argmax}{arg\,max}
\DeclareMathOperator*{\argmin}{arg\,min}
\newcommand{\circled}[1]{\tikz[baseline=(char.base)]{\node[shape=circle,draw,inner sep=1pt] (char) {#1};}}
\acmConference[CIKM '25]{Proceedings of the 34th
ACM International Conference on Information and Knowledge
Management}{November 10--14, 2025}{Seoul, Republic of Korea}
\acmBooktitle{Proceedings of the 34th ACM International Conference on
Information and Knowledge Management (CIKM '25), November 10--14, 2025,
Seoul, Republic of Korea}
\acmDOI{10.1145/3746252}

\acmISBN{979-8-4007-2040-6/2025/11}

\begin{document}

\title{ADMP-GNN: Adaptive Depth Message Passing GNN}

\author{Yassine Abbahaddou}

\affiliation{%
  \institution{LIX, Ecole Polytechnique, IP Paris}
  \city{Palaiseau}
  \country{France}
}
\email{yassine.abbahaddou@polytechnique.edu}

\author{Fragkiskos D. Malliaros}
\affiliation{%
  \institution{Université Paris-Saclay, CentraleSupélec, Inria}
  \city{Gif-sur-Yvette}
  \country{France}
  }
\email{fragkiskos.malliaros@centralesupelec.fr}

\author{Johannes F.	Lutzeyer}
\affiliation{%
  \institution{LIX, Ecole Polytechnique, IP Paris}
  \city{Palaiseau}
  \country{France}
}
\email{johannes.lutzeyer@polytechnique.edu}

\author{Michalis Vazirgiannis}
\affiliation{%
 \institution{LIX, Ecole Polytechnique, IP Paris}
  \city{Palaiseau}
  \country{France}
 }
 \email{mvazirg@lix.polytechnique.fr}

\renewcommand{\shortauthors}{Yassine	Abbahaddou, Fragkiskos D. Malliaros, Johannes F. Lutzeyer, \& Michalis Vazirgiannis}

\begin{abstract}
 Graph Neural Networks (GNNs) have proven to be highly effective in various graph learning tasks. A key characteristic of GNNs is their use of a fixed number of message-passing steps for all nodes in the graph, regardless of each node's diverse computational needs and characteristics.  Through empirical real-world data analysis, we demonstrate that the optimal number of message-passing layers varies for nodes with different characteristics. This finding is further supported by experiments conducted on synthetic datasets. To address this, we propose \emph{Adaptive Depth Message Passing  GNN (ADMP-GNN)}, a novel framework that dynamically adjusts the number of message passing layers for each node, resulting in improved performance. This approach applies to any model that follows the message passing scheme. We evaluate ADMP-GNN on the node classification task and observe performance improvements over baseline GNN models. Our code is publicly available at: \href{https://github.com/abbahaddou/ADMP-GNN}{https://github.com/abbahaddou/ADMP-GNN}.
 \end{abstract}





\maketitle

\section{Introduction}
A plethora of structured data comes in the form of graphs \cite{bornholdt2001handbook,MALLIAROS201395,cao2020spectral}. This has driven the need to develop neural network models, known as Graph Neural Networks (GNNs), that can effectively process and analyze graph-structured data. GNNs have garnered significant attention for their ability to learn complex node and graph representations, achieving remarkable success in several practical applications \cite{rampavsek2022recipe,corso2022diffdock,glie-asonam24,pmlr-v202-duval23a,castro-correa-tnnls24}. Many of these models are instances of Message Passing Neural Networks (MPNNs) \cite{gilmer2017neural,xu2019powerful,Kipf:2017tc}. A common characteristic of GNNs is that they typically employ a fixed number of message passing steps for all nodes, determined by the number of layers in the GNN \cite{gilmer2017neural}. This static framework raises an intriguing question: \emph{Should the number of message passing steps be adapted individually for each node to better capture their unique characteristics and computational needs?}

Determining the optimal number of message passing layers for each node in a GNN presents a significant challenge due to the intricate and diverse nature of graph structures, node features, and learning tasks. While deeper GNNs can capture long-range dependencies \cite{liu2021eignn}, they can also encounter issues like oversmoothing, where nodes become indistinguishably similar \cite{luan2022revisiting,giraldo2023sjlr}.  In dense graphs, where information can propagate quickly, even shallow GNNs can effectively capture local information \cite{zeng2020deep}. Conversely, sparse graphs may require additional layers to facilitate effective information sharing \cite{zhang2021evaluating,Zhao2020PairNorm}. This underscores the importance of selecting the appropriate number of layers for a GNN to capture the necessary graph information effectively. An even more compelling idea is to adjust the GNN depth for each node based on its local structural properties. This adaptive approach could be especially beneficial for graphs with varied local structures, ensuring that each node is processed according to its unique requirements.

This need for per-node customization naturally aligns with the concept of \emph{Dynamic Neural Networks}, also referred to as \emph{Adaptive Neural Networks} \cite{bolukbasi2017adaptive}. Dynamic Neural Networks represent a class of models that can adjust their architecture or parameters depending on the input. Dynamic Neural Networks have gained significant popularity, especially in the field of computer vision. Examples of adaptation include varying the number of layers and implementing skip connections  \cite{li2017pruning,huang2016deep,sabour2017dynamic}. Beyond computer vision, other types of adaptive neural networks have been explored in various domains. In natural language processing, adaptive computation time models allow recurrent neural networks (RNNs) to determine the required number of recurrent steps dynamically based on the complexity of the input sequence \cite{graves2016adaptive}. However, applying these adaptations to graph learning tasks presents unique challenges. Unlike the structured and homogeneous data often encountered in computer vision, graph data involves overcoming nuisances related to the inherent complex structure of graphs. Graphs may have varying node degrees, non-uniform connectivity, and feature Heterogeneity. While some dynamic approaches may extend effectively to graph classification tasks, applying these methods to node classification presents additional complexities. In node classification, each input sample (node) is part of a larger interconnected system where information propagates through edges, i.e., the prediction for one node often depends on the features of the other nodes and the structure of the graph. As a result, specific techniques are needed to account for the dependencies and relational information encoded in the graph structure.

In this work, we focus on the task of node classification by proposing ADMP-GNN, a novel approach that dynamically adapts the number of layers for each node within a GNN. Our main contributions are as follows:
\begin{itemize}
    \item \textbf{Node-specific depth analysis in GNNs.} We demonstrate through empirical analysis that different nodes within the same graph may require varying numbers of message passing steps to accurately predict their labels. This finding underscores the importance of node-specific depth in GNNs.
    \item \textbf{Adaptive message passing layer integration.} We present ADMP-GNN, a novel approach that enables any GNN to make predictions for each node at every layer. Training the GNN to predict labels across all layers is a multi-task setting, which often suffers from gradient conflicts, leading to suboptimal performance. To address this, we propose a sequential training methodology where layers are progressively trained, and their gradients are subsequently frozen, thereby mitigating conflicts and improving overall performance.
    \item \textbf{Adaptive layer policy learning for node classification.} We introduce a heuristic method to learn a layer selection policy using a set of validation nodes. This policy is then applied to select the optimal layer for predicting the labels of test nodes, ensuring that each node exits the GNN at the most appropriate layer for its specific classification task.
    \item \textbf{Model-agnostic flexibility.} Our approach is model-agnostic and can be integrated with any GNN architecture that employs a message passing scheme. This flexibility enhances the GNN's performance on node classification tasks, providing a significant improvement over traditional fixed-layer approaches.
    
\end{itemize}

\section{Related Work}
In this section, we review key developments in GNNs and dynamical neural networks, which form the foundation of our work. 

\subsection{Graph Neural Networks}
Graphs, denoted as $\mathcal{G} = (\mathcal{V}, \mathcal{E}, X)$, are mathematical structures used to represent pairwise relationships. Here, $\mathcal{V}$ is the set of nodes, $\mathcal{E}$ is the set of edges, and $X = \{x_u : u \in \mathcal{V}\}$  represents the node features, where each $x_u \in \mathbb{R}^d$ corresponds to a $d$-dimensional feature vector associated with node $u$. GNNs are a class of deep learning methods specifically designed to operate on such graph-structured data. As with most neural networks, GNNs are formed by stacking many layers.  Each layer, $\ell$, is responsible for updating the node representations $\{h^{(\ell)}_u: u \in \mathcal{V}\}_{0\leq \ell \leq L}$, relying on the graph structure and the output from the previous layer, $\{h^{(\ell-1)}_u:u \in \mathcal{V}\}$. Conventionally, the features of the nodes are used as input of the first layer, i.e., $h^{(0)}_u = x_u \in \mathbb{R}^{d}$ for all $u\in \mathcal{V}$. A basic GNN layer is based on the message passing mechanism and consists of two components: (i)~\textit{Aggregate Layer} $\psi$ that applies for each node $v$,  a permutation invariant function to its neighbors, denoted by $\mathcal{N}(v) $ to generate the aggregated node feature;  (ii)~\textit{Update Layer} $\phi$ that combines the aggregated node feature $m^{(\ell)}_v$ with the previous hidden vector $h^{(\ell-1)}_v$, and generate a new representation $h^{(\ell)}_v$ of the same node $v$,
\begin{align*} 
   m^{(\ell)}_v&= \psi^{(\ell)}(\{h^{(\ell-1)}_u: u \in \mathcal{N}(v)\}), \\ 
   h^{(\ell)}_v&=\phi^{(\ell)}(h^{(\ell-1)}_v , m^{(\ell)}_v).
\end{align*}
Depending on the task, an additional readout or pooling function can be added after the last layer to aggregate the representation of nodes, as follows, $h_G = \text{ReadOut}(\{h^{(L)}_u: u \in \mathcal{V}\})$. Examples of the most popular GNN models include the Graph Convolutional Network (GCN) \cite{Kipf:2017tc} and Graph Isomorphism Network (GIN) \cite{xu2019powerful}, which apply a weighted sum and sum operator in their aggregate layer, respectively.

\subsection{Dynamical Neural Networks}
Dynamic neural networks are gaining significance in the field of deep learning. Unlike static models with fixed computational graphs and parameters during inference, dynamic networks adapt their structures or parameters based on varying inputs. This dynamic flexibility gives models significant benefits, such as improved accuracy, enhanced computational efficiency, and superior adaptability \cite{DBLP:journals/corr/abs-2010-05300,DBLP:journals/corr/abs-2006-04152}. A popular type of dynamic neural networks includes those that dynamically adjust network depth based on each input. For instance, in natural language processing, some adaptive large language models employ adaptive depth to optimize both inference speed and computational memory usage of the Transformer architecture \cite{Elbayad2020Depth-Adaptive,schuster2022confident,DBLP:journals/corr/VaswaniSPUJGKP17}. In the field of computer vision, there are studies that dynamically generate filters conditioned on each input, enhancing flexibility without significantly increasing the number of model parameters \cite{jia2016dynamic}.

In Graph Machine Learning, dynamic adaptations have been proposed for GNN message passing. These methods include dynamically determining which neighbors to consider at each layer, enabling more flexible and adaptive message passing \cite{finkelshtein2023cooperative}, or allowing nodes to react to individual messages at varying times rather than processing aggregated neighborhood information synchronously \cite{faber2024gwac}. Another line of work focuses on adapting normalization layers for each node to enhance expressive power, generating representations that reflect local neighborhood structures \cite{eliasof2024granola}. Additionally, other related approaches use residual connections to mitigate issues like oversmoothing \cite{errica2023adaptive,DBLP:journals/corr/abs-2007-02133}. In \cite{spinelli2020adaptive} the authors propose the AP-GCN model, which includes a node-specific halting unit that is intended to learn the node-wise optimal message passing depth. To the best of our knowledge, besides \cite{spinelli2020adaptive}, in the field of GNNs, there has been no prior work proposing adaptive depth for each node. However, several studies have focused on combining all GNN layers. These works typically aim to adapt GNN architectures for  heterophilic graphs \cite{chien2020adaptive} and leverage information from higher-order neighbors \cite{DBLP:journals/corr/abs-1806-03536}. While combining GNN layers can be viewed as a form of depth-adaptive strategy, where the final node representation is guided by the optimal intermediate hidden states, this approach remains \emph{static} because the same inference policy is applied uniformly across all nodes and learned layer aggregators stay fixed after training.

\section{Adaptive Depth Message Passing GNN} \label{method}
\begin{figure}[t]
    \centering
    \includegraphics[width=\linewidth]{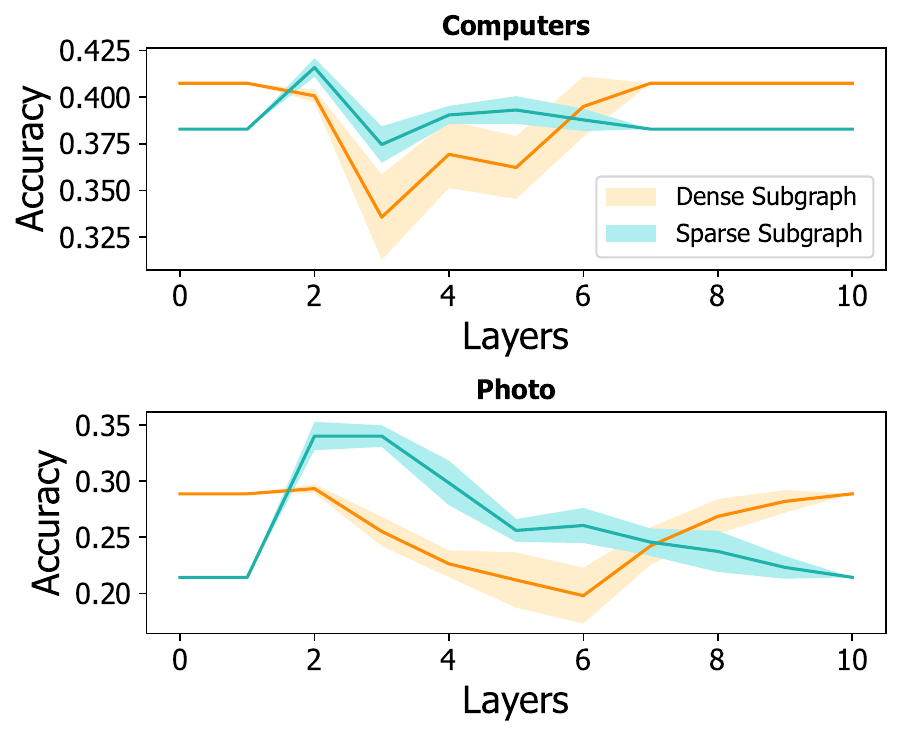}
    \caption{Effect of GCN's depth on sparse and dense subgraphs. The figure shows the performance of GCNs when varying layer depths, and comparing its effectiveness on both sparse and dense subgraphs.}
    \label{fig:photo_synthetic}
\end{figure}

In this section, we first present an empirical analysis highlighting the necessity for node-specific depth in GNNs. Then, we introduce our ADMP-GNN. Our study includes experiments on both synthetic and real-world datasets to illustrate the importance and potential benefits of this methodology. 

\subsection{Depth Analysis on Synthetic Graphs}
This analysis aims to highlight the importance of employing a varying number of message passing steps based on the specific characteristics of individual nodes. As outlined in the introduction, this approach is particularly relevant for graphs where nodes exhibit diverse properties, such as varying local structures. In this experiment, we investigate the effect of the number of message-passing layers on nodes with varied levels of local neighborhood sparsity. To achieve this, we construct a graph by merging two subgraphs extracted from real-world datasets, namely Computers and Photo \cite{cs_data}. Both subgraphs contain the same number of nodes, exhibit nearly identical homophily, and have equally distributed node labels. The main difference between these subgraphs lies in their structure, as one subgraph is sparse while the other is dense. Consequently, nodes within each subgraph share comparable structural characteristics. Details on the construction of these synthetic datasets and visualizations of their adjacency matrices are provided in Appendix \ref{app:motivation}.

We have trained $L+1$ different GCN models, with a varying number of layers $\ell\in \{ 0,\ldots,L\}$, where $L$ represents the maximum depth, set to $L=10$. Each GCN was trained on the entire synthetic graph, composed of both sparse and dense subgraphs, but the performance was evaluated separately on the individual subgraphs. This allows us to assess the impact of GNN depth on different types of local subgraph structures. The results, presented in Fig. \ref{fig:photo_synthetic}, reveal notable differences in behavior between the subgraph types.
As observed, in dense subgraphs, the test accuracy decreases at a faster rate, while in sparse subgraphs, the drop in accuracy occurs later, typically around layers 2 or 3. Moreover, the optimal number of layers differs between sparse and dense subgraphs. For instance, in the Computers dataset, the highest accuracy is achieved at layer 2 for the sparse subgraph, while for the dense subgraph, the optimal performance is reached at layer 0. Additionally, in the Photo dataset, we observe a distinct behavior starting from layer 6, where the impact of GNN depth diverges between sparse and dense subgraphs. This highlights the need to adapt the number of layers per node based on its characteristics.

\subsection{Adaptive Message Passing Layer Integration}
Building on the insights from the previous analysis, it is essential to develop a framework that can efficiently adjust the GNN depth. For a maximum GNN depth $L$, a traditional approach would involve training $L+1$ different GNNs, each with a distinct number of layers $\ell$, where $\ell$ ranges from 0 and $L$. Subsequently, a policy must be established to determine the optimal GNN with the appropriate number of layers for each individual node. However, training $L+1$ GNNs separately can be computationally expensive. A more efficient approach involves designing a single GNN with $L+1$ layers that provide predictions at each intermediate layer. To ensure equivalence to the previous approach, i.e., training $L+1$ GNNs separately, the computational graph for predictions at layer $\ell$ must match that of a standard GNN with $\ell$ layers, and the classification performance at layer $\ell$ should yield results comparable to those of a conventional GNN with $\ell$ layers. 
\begin{figure*}
    \centering
    \includegraphics[width=0.8\textwidth]{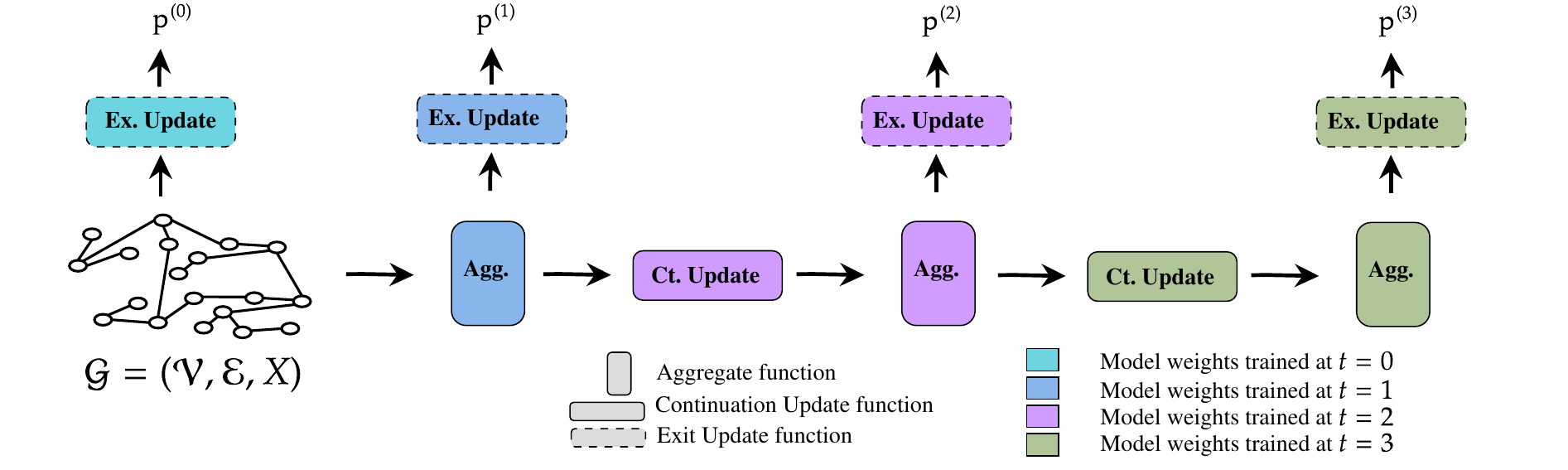}
    \caption{Illustration of ADMP-GNN, when the maximum GNN depth is $L=3$.}
    \label{fig:AMP_architerchture}
\end{figure*}

To address these challenges, we introduce ADMP-GNN, an adaptation of a Message Passing Neural Network with a maximum depth of $L$ layers.  The goal is to ensure that the computational graph and the performance of ADMP-GNN at a certain layer match that of traditional GNNs when trained and tested on the same number of layers. To achieve this, we incorporate an additional \textit{Update} function, denoted as $\phi^{(\ell)}_{\mathbf{Ex}}$ ($\mathbf{Ex}$ stands for `Exit'), to directly predict node labels at a given layer $\ell$, i.e., $p^{(\ell)}$. The function $\phi^{(\ell)}_{\mathbf{Ex}}$ is defined as follows,
\begin{equation*}
    p^{(\ell)}_v = \phi^{(\ell)}_{\mathbf{Ex}} \left (   h^{(\ell-1)}_v , m^{(\ell)}_v \right ) \\
             = \text{Softmax} \left (   \widetilde{W}^{(\ell)} m^{(\ell)}_v \right ),
\end{equation*}
where $\widetilde{W}^{(\ell)} \in \mathbb{R}^{d^{(\ell)}\times c}$ is a learnable weight matrix, $d^{(\ell)}$ is the dimension of the hidden representation at the $\ell$-th layer, and $c$ is the number of classes. To obtain predictions at a deeper layer $\ell^\prime  \geq \ell $, we continue the message passing using another \textit{Update} function $\phi^{(\ell)}_{\mathbf{Ct}}$ ($\mathbf{Ct}$ stands for `Continuation'),
    \begin{align*}
        p^{(\ell)}_v  &= \phi^{(\ell)}_{\mathbf{Ex}} \left (   h^{(\ell-1)}_v , m^{(\ell)}_v \right ),\\
        h^{(\ell+1)}_v &= \phi^{(\ell)}_{\mathbf{Ct}}\left (  h^{(\ell-1)}_v, m^{(\ell)}_v\right ).
    \end{align*}
For $\ell=0$, we directly use the \textit{Exit Update} function on the node features, i.e.,  $m^{(0)}_v = x_v$,
$$\forall v \in \mathcal{V}, \quad p_v^{(0)} =  \phi^{(0)}_{\mathbf{Ex}} \left ( m^{(0)}_v \right ) \\
              = \text{Softmax} \left (   \widetilde{W}^{(0)} x_v \right ).$$
In Fig. \ref{fig:AMP_architerchture}, we illustrate the architecture of the proposed ADMP-GNN.
\subsection{Training Scheme of ADMP-GNN}
Our next objective is to train ADMP-GNN to predict node labels across all layers $ \ell \in \{ 0,\ldots , L\}$ simultaneously. For each layer $\ell$, let $\theta_\ell$ denote the weights of the function $\psi ^{(\ell)}\circ\phi_{\mathbf{Ct}}^{(\ell)}(\cdot)$. Two distinct strategies for training ADMP-GNN are explored:

\paragraph{1. Aggregate Loss Minimization (ALM)} This straightforward approach minimizes the aggregate loss over all layers. The total loss is formulated as,
\begin{align}
\mathcal{L}_{\text{ALM}} & := 
    \argmin_{\theta} \mathbb{E}_{v\in \mathcal{V}}\left [ S_L(v) \right ]\label{eq:multi_obj}\\ 
    & =\argmin_{\theta_0, \ldots, \theta_{L}} \mathbb{E}_{v\in \mathcal{V}}\left [ \sum_{\ell=0}^{L}\mathcal{L}\left ( p_v^{(\ell)}(m^{(0)}_v, \theta_0, \ldots, \theta_{\ell}) , y_v\right ) \right ],\nonumber
\end{align}
where  $p_v^{(\ell)}(m^{(0)}_v, \theta_0, \ldots, \theta_{\ell}) = \phi^{(\ell)}_{\mathbf{Ex}}(m^{(\ell)}_v)$ is the prediction for node $v$ at  layer $\ell$, and $\mathcal{L}$ is the Cross Entropy Loss. This approach may encounter gradient conflicts, particularly for early layers involved in both computation and back-propagation across upper layers. 

\paragraph{2. Sequential Training (ST)} We have studied an alternative training setup where we progressively train one GNN layer at a time, subsequently freezing each layer after training. More formally, the problem in \eqref{eq:multi_obj} can be tackled using dynamic programming as follows,
\begin{align*}
    \forall v\in \mathcal{V}, S_{\ell+1}(v) &= \mathcal{L}\left ( p_v^{(\ell)}(m^{(0)}_v,\theta_0^\star, \ldots, \theta_{\ell}^\star, \theta_{\ell+1}) , y_v\right ) \\
    & ~~~~~~~~~~~~~~~~~~~~~~~~~~~~~~~~~~~~~~~~+  S_\ell (v)\\
    \theta_{\ell}^\star&  = \argmin_{\theta_{\ell}} \mathbb{E}_{v\in V}\left [S_\ell(v)\right ],
\end{align*}
where $\forall v \in \mathcal{V}, ~ S_0(v) = \mathcal{L}(\phi_{\mathbf{Ex}}^{(0)}, y_v)$. For each intermediate layer $\ell<L$, by training this layer on the node classification task, we obtain high-quality node representations $\{h^{(\ell)}_v:v \in \mathcal{V}\}$. These representations are directly employed for predictions and serve as a robust foundation for the label predictions of the subsequent layer $\ell+1$. Algorithm \ref{algo:seq_training} offers a summary of the approach.

\begin{algorithm}[t]
\caption{Sequential Training for ADMP-GNN}\label{algo:seq_training}

\begin{algorithmic}
\State \textbf{Inputs: } Graph $\mathcal{G}=(\mathcal{V},\mathcal{E},X)$, number of layers $L$, node classification loss function $\mathcal{L},$\\
\For{$t=0$ \textbf{to} $L$}
  \If{$t=0$}
    \State 1. Set $ h^{(0)}_v \gets m^{(0)}_v = x_v$ for all $ v \in \mathcal{V}.$
    \State 2. Compute predictions at layer $\ell=0$, i.e., $$\forall v \in \mathcal{V}, \quad p_v^{(0)} =  \phi^{(0)}_{\mathbf{Ex}} ( h^{(0)}_v  ).$$
    \State 3. Train the weights of $\phi^{(0)}_{\mathbf{Ex}}$ to minimize $\mathcal{L}(p_v^{(0)}).$
    \State 4. Freeze the gradients of $\phi^{(0)}_{\mathbf{Ex}}.$
  \Else
    \State 1. Use the $\phi^{(t-1)}_{\mathbf{Ct}}$ to update node representations 
        $$\forall v \in \mathcal{V}, \quad \tilde{h}^{(t-1)}_v =  \phi^{(t-1)}_{\mathbf{Ct}}(h^{(t-1)}_v). $$
    \State 2. Aggregate the information for neighbor nodes
        $$\forall v \in \mathcal{V}, \quad m^{(t)}_v = \psi^{(t)}(\{ \tilde{h}^{(t-1)}_u: u \in \mathcal{N}(v)\}). $$
    \State 3. Compute predictions at layer $\ell=t$, i.e., $$\forall v \in \mathcal{V}, \quad p_v^{(t)} = \phi^{(t-1)}_{\mathbf{Ct}}(  \tilde{h}^{(t-1)}_v, m^{(t)}_v).$$
    \State 4. Train $\phi^{(t-1)}_{\mathbf{Ct}}, \psi^{(t)}, \text{ and } \phi^{(t)}_{\mathbf{Ex}}$ to minimize $\mathcal{L}(p_v^{(t)}).$
    \State 5. Freeze the gradients of $\phi^{(t-1)}_{\mathbf{Ct}}, \psi^{(t)}, \text{ and } \phi^{(t)}_{\mathbf{Ex}}.$ 
  \EndIf
\EndFor
\end{algorithmic}
\end{algorithm}

 \begin{table*}[t]
\caption{Comparison of ADMP-GCN training paradigms ALM and ST. These paradigms are also compared to the single-task training setting to evaluate which approach most closely mimics the classical GCN under single-task training.  The best multi-task results for each dataset are \textbf{bolded}.}
\resizebox{0.8\textwidth}{!}{%
\begin{tabular}{c|clcccccc}
\toprule
\# Layers & Training Paradigm & Model & Cora & CiteSeer & CS & PubMed & Genius & Ogbn-arxiv \\
\midrule
\multirow{3}{*}{\circled{0}} & \multirow{1}{*}{Single-task} & GCN & $56.38 {\scriptstyle \pm 0.04}$ & $57.18 {\scriptstyle \pm 0.12}$ & $88.04 {\scriptstyle \pm 0.49}$ & $72.50 {\scriptstyle \pm 0.09}$ & $80.82 {\scriptstyle \pm 1.00}$ & $48.88 {\scriptstyle \pm 0.06}$ \\
\cmidrule(lr){2-9}
& \multirow{2}{*}{Multi-task} & ADMP-GCN (ALM) & $\mathbf{56.96 {\scriptstyle \pm 0.20}}$ & $\mathbf{58.44 {\scriptstyle \pm 0.21}}$ & $87.06 {\scriptstyle \pm 1.06}$ & $72.11 {\scriptstyle \pm 0.18}$ & $80.03 {\scriptstyle \pm 0.37}$ & $36.50 {\scriptstyle \pm 0.12}$ \\
& & ADMP-GCN (ST) & $56.38 {\scriptstyle \pm 0.06}$ & $57.17 {\scriptstyle \pm 0.09}$ & $\mathbf{87.27 {\scriptstyle \pm 1.29}}$ & $\mathbf{72.48 {\scriptstyle \pm 0.14}}$ & $\mathbf{80.17 {\scriptstyle \pm 0.79}}$ & $\mathbf{48.86 {\scriptstyle \pm 0.03}}$ \\
\midrule
\multirow{3}{*}{\circled{1}} & \multirow{1}{*}{Single-task} & GCN & $76.90 {\scriptstyle \pm 0.14}$ & $69.68 {\scriptstyle \pm 0.06}$ & $91.74 {\scriptstyle \pm 0.80}$ & $76.63 {\scriptstyle \pm 0.13}$ & $80.23 {\scriptstyle \pm 0.37}$ & $55.21 {\scriptstyle \pm 0.50}$ \\
\cmidrule(lr){2-9}
& \multirow{2}{*}{Multi-task} & ADMP-GCN (ALM) & $75.67 {\scriptstyle \pm 0.18}$ & $\mathbf{70.12 {\scriptstyle \pm 0.04}}$ & $90.55 {\scriptstyle \pm 0.74}$ & $73.74 {\scriptstyle \pm 0.16}$ & $\mathbf{80.13 {\scriptstyle \pm 0.29}}$ & $39.54 {\scriptstyle \pm 1.44}$ \\
& & ADMP-GCN (ST) & $\mathbf{76.90 {\scriptstyle \pm 0.00}}$ & $69.70 {\scriptstyle \pm 0.00}$ & $\mathbf{90.89 {\scriptstyle \pm 0.81}}$ & $\mathbf{76.60 {\scriptstyle \pm 0.00}}$ & $79.93 {\scriptstyle \pm 0.00}$ & $\mathbf{55.15 {\scriptstyle \pm 0.00}}$ \\
\midrule
\multirow{3}{*}{\circled{2}} & \multirow{1}{*}{Single-task} & GCN & $81.06 {\scriptstyle \pm 0.50}$ & $71.05 {\scriptstyle \pm 0.48}$ & $91.67 {\scriptstyle \pm 0.94}$ & $79.46 {\scriptstyle \pm 0.31}$ & $79.88 {\scriptstyle \pm 0.51}$ & $66.92 {\scriptstyle \pm 0.67}$ \\
\cmidrule(lr){2-9}
& \multirow{2}{*}{Multi-task} & ADMP-GCN (ALM) & $70.82 {\scriptstyle \pm 4.05}$ & $60.95 {\scriptstyle \pm 2.39}$ & $31.20 {\scriptstyle \pm 12.85}$ & $75.10 {\scriptstyle \pm 2.41}$ & $79.64 {\scriptstyle \pm 0.60}$ & $55.25 {\scriptstyle \pm 0.92}$ \\
& & ADMP-GCN (ST) & $\mathbf{80.73 {\scriptstyle \pm 0.33}}$ & $\mathbf{71.33 {\scriptstyle \pm 0.40}}$ & $\mathbf{91.49 {\scriptstyle \pm 0.66}}$ & $\mathbf{79.02 {\scriptstyle \pm 0.21}}$ & $\mathbf{80.06 {\scriptstyle \pm 0.11}}$ & $\mathbf{66.51 {\scriptstyle \pm 0.65}}$ \\
\midrule
\multirow{3}{*}{\circled{3}} & \multirow{1}{*}{Single-task} & GCN & $79.14 {\scriptstyle \pm 1.58}$ & $66.33 {\scriptstyle \pm 1.35}$ & $89.80 {\scriptstyle \pm 0.87}$ & $78.50 {\scriptstyle \pm 0.68}$ & $80.00 {\scriptstyle \pm 0.04}$ & $67.33 {\scriptstyle \pm 0.55}$ \\
\cmidrule(lr){2-9}
& \multirow{2}{*}{Multi-task} & ADMP-GCN (ALM) & $68.64 {\scriptstyle \pm 5.14}$ & $51.30 {\scriptstyle \pm 6.74}$ & $47.82 {\scriptstyle \pm 12.98}$ & $74.17 {\scriptstyle \pm 1.98}$ & $\mathbf{80.04 {\scriptstyle \pm 0.10}}$ & $56.22 {\scriptstyle \pm 0.50}$ \\
& & ADMP-GCN (ST) & $\mathbf{80.21 {\scriptstyle \pm 0.52}}$ & $\mathbf{70.08 {\scriptstyle \pm 0.90}}$ & $\mathbf{89.83 {\scriptstyle \pm 1.02}}$ & $\mathbf{78.25 {\scriptstyle \pm 0.51}}$ & $79.93 {\scriptstyle \pm 0.00}$ & $\mathbf{68.31 {\scriptstyle \pm 0.48}}$ \\
\midrule
\multirow{3}{*}{\circled{4}} & \multirow{1}{*}{Single-task} & GCN & $75.96 {\scriptstyle \pm 1.93}$ & $60.33 {\scriptstyle \pm 2.38}$ & $78.90 {\scriptstyle \pm 22.33}$ & $76.59 {\scriptstyle \pm 0.98}$ & $80.01 {\scriptstyle \pm 0.04}$ & $65.49 {\scriptstyle \pm 0.99}$ \\
\cmidrule(lr){2-9}
& \multirow{2}{*}{Multi-task} & ADMP-GCN (ALM) & $67.88 {\scriptstyle \pm 6.05}$ & $49.29 {\scriptstyle \pm 7.87}$ & $52.76 {\scriptstyle \pm 11.79}$ & $73.40 {\scriptstyle \pm 1.95}$ & $\mathbf{80.04 {\scriptstyle \pm 0.10}}$ & $56.43 {\scriptstyle \pm 0.31}$ \\
& & ADMP-GCN (ST) & $\mathbf{81.05 {\scriptstyle \pm 0.49}}$ & $\mathbf{67.99 {\scriptstyle \pm 0.74}}$ & $\mathbf{88.86 {\scriptstyle \pm 0.70}}$ & $\mathbf{75.27 {\scriptstyle \pm 1.02}}$ & $79.93 {\scriptstyle \pm 0.00}$ & $\mathbf{69.29 {\scriptstyle \pm 0.74}}$ \\
\midrule
\multirow{3}{*}{\circled{5}} & \multirow{1}{*}{Single-task} & GCN & $70.09 {\scriptstyle \pm 4.01}$ & $57.40 {\scriptstyle \pm 3.43}$ & $77.96 {\scriptstyle \pm 15.24}$ & $74.32 {\scriptstyle \pm 3.66}$ & $80.01 {\scriptstyle \pm 0.04}$ & $63.04 {\scriptstyle \pm 1.33}$ \\
\cmidrule(lr){2-9}
& \multirow{2}{*}{Multi-task} & ADMP-GCN (ALM) & $67.68 {\scriptstyle \pm 7.04}$ & $50.14 {\scriptstyle \pm 7.65}$ & $54.53 {\scriptstyle \pm 10.61}$ & $73.27 {\scriptstyle \pm 1.51}$ & $\mathbf{80.04 {\scriptstyle \pm 0.10}}$ & $56.66 {\scriptstyle \pm 0.31}$ \\
& & ADMP-GCN (ST) & $\mathbf{81.22 {\scriptstyle \pm 0.36}}$ & $\mathbf{67.42 {\scriptstyle \pm 0.75}}$ & $\mathbf{86.65 {\scriptstyle \pm 1.52}}$ & $\mathbf{75.08 {\scriptstyle \pm 0.94}}$ & $79.93 {\scriptstyle \pm 0.00}$ & $\mathbf{68.92 {\scriptstyle \pm 1.00}}$ \\
\bottomrule
\end{tabular}}
\label{tab:multi_task_updated}
\end{table*}

\paragraph{Comparison of ADMP-GNN Training Paradigms} To identify the optimal multi-task training configuration, we evaluate how much of a performance drop we incur at each layer compared to the single-task setting in GNNs. The comparative analysis of the three strategies for both GCN is detailed in Tables \ref{tab:multi_task_updated} and \ref{tab:multi_task_GIN}. Our findings indicate that ADMP-GNN ST outperforms ADMP-GNN ALM. Notably, the performance of ADMP-GNN ST is comparable to, or even exceeds, that of GNN when trained under the single-task setting. Furthermore, ADMP-GNN ST exhibits a smaller standard deviation, suggesting more consistent performance. These results underscore the effectiveness of the ST setting in addressing gradient conflicts inherent in aggregate loss minimization (ALM). Furthermore, ADMP-GNN ST effectively mimics the results of training $L+1$ separate GNNs, each with a different number of layers, while requiring only a single unified model. The next step, outlined in Section \ref{subsec:policy}, is to learn a policy that selects the optimal prediction layer for each node, completing the framework of ADMP-GNN.

\paragraph{Time Complexity.} The training setup ST, where we sequentially train the deep ADMP-GNN, incurs relatively higher time costs due to the need for $L+1$ training iterations. However, in each iteration, backpropagation is performed on a limited number of parameters, approximately equivalent to those in a single message passing layer. Consequently, only a small number of epochs are required for each training iteration. We report the training time of each approach in Table \ref{tab:time_complexity} in Appendix \ref{app:time_comp}.

\subsection{Empirical Insights into Node Specific Depth}
We define \textit{Oracle Accuracy} as the maximum achievable test accuracy under an optimal policy for selecting layers. This is formally expressed as,
$$\mathcal{A}_{\text{oracle}} = \frac{1}{|\mathcal{V}_{\text{test}}|} \sum_{v \in \mathcal{V}_{\text{test}}} \mathbbm{1}\left( y_v \in  \{\widehat{y}_v^{(\ell)}: ~~\ell = 0, \ldots , L\}  \right),$$
where $\mathcal{V}_{\text{test}}$ is the set of test nodes, $\{\widehat{y}_v^{(\ell)}: \ell = 0, \ldots , L\}$ represents the predictions for node $v$ at each layer, and $\mathbbm{1}(\cdot)$ is the indicator function. Specifically, $\mathcal{A}_{\text{oracle}}$ corresponds to the accuracy obtained if, for each test node, we could perfectly choose the layer that provides the correct prediction. Assuming such an optimal selection for each node, it represents the upper bound of accuracy achievable when employing adaptive strategies to determine the best layer for each node.

Based on the findings presented in Table \ref{tab:oracle}  for GCN, it is clear that $\mathcal{A}_{\text{oracle}}$ outperforms the highest accuracies achieved by both GCN and ADMP-GCN ALM. This empirical evidence strongly indicates that adopting a distinct exit layer for each node is highly beneficial in an optimal configuration.

\begin{table*}[ht]
\centering
\caption{Comparison of highest accuracy ($\pm$ standard deviation) for GCN and ADMP-GCN ST, with the layer achieving the best accuracy indicated in brackets. The final row shows the \textit{Oracle Accuracy} for ADMP-GCN ST. Best results for each dataset are \textbf{bolded}.}
\resizebox{0.8\textwidth}{!}{%
\begin{tabular}{lllllll}
\toprule
 Model & Cora & CiteSeer & CS & PubMed & Genius & Ogbn-arxiv \\  \midrule
 GCN & $81.06 {\scriptstyle \pm 0.50} ~[2]$  &  $71.05 {\scriptstyle \pm 0.48} ~[2]$ & $91.67 {\scriptstyle \pm 0.94} ~[2]$  &  $79.46 {\scriptstyle \pm 0.31} ~[2]$ & $80.82 {\scriptstyle \pm 1.00} ~[0]$ &  $67.33 {\scriptstyle \pm 0.55} ~[3]$ \\

ADMP-GCN (ST) & $81.22 {\scriptstyle \pm 0.36} ~[5]$ & $71.33 {\scriptstyle \pm 0.40} ~[2]$ & $91.49 {\scriptstyle \pm 0.66} ~[2]$ & $79.02 {\scriptstyle \pm 0.21} ~[2]$ & $80.17 {\scriptstyle \pm 0.79} ~[0]$ & $69.29 {\scriptstyle \pm 0.74} ~[4]$ \\
ADMP-GCN (ST) - Oracle & $\mathbf{89.43 {\scriptstyle \pm 0.19}}$ & $\mathbf{81.96 {\scriptstyle \pm 0.49}}$ & $\mathbf{97.24 {\scriptstyle \pm 0.52}}$ & $\mathbf{90.13 {\scriptstyle \pm 0.36}}$ & $\mathbf{85.97 {\scriptstyle \pm 7.59}}$ & $\mathbf{79.64 {\scriptstyle \pm 0.27}}$ \\ \bottomrule

\end{tabular}
}

\label{tab:oracle}
\end{table*}

\subsection{Generalization to Test Nodes}\label{subsec:policy}
In this section, we introduce a heuristic approach to predict the optimal exit layer for test nodes based on the assumption that nodes exhibiting \textit{structural similarity} should share the same exit layer. The notion of structural similarity can be assessed using various metrics. In our work, we define the structural similarity based on node centrality metrics, i.e., nodes are considered structurally similar if they exhibit closely aligned centrality values within the graph. We measured node centralities using both local metrics like degree and global metrics such as $k$-core \cite{kcore-vldbj20}, PageRank scores \cite{brin1998anatomy}, and Walk Count, which are detailed below.

\paragraph{$k$-core} The $k$-core decomposition of a graph is a method that involves systematically pruning nodes. In each iteration, a subgraph $\mathcal{G}_k \subset \mathcal{G}$ is constructed by removing all vertices whose degree is less than $k$. The core number of a vertex $u$, denoted as $\text{core}(u)$, represents the largest value of $k$ for which $i$ is included in the $k$-core. Formally, this is expressed as $\text{core}(i) = \max \{ k : i \in \mathcal{G}_k \}$. Nodes with higher core numbers are typically found in denser subgraphs, reflecting strong interconnections with their neighbors and indicating a higher level of centrality.

\paragraph{PageRank} A node’s PageRank score represents the probability of a random walk visiting that node, making it a fundamental metric for measuring node importance in social network analysis and the web \cite{brin1998anatomy}.

\paragraph{Walk Count} The Walk Count centrality quantifies the importance of a node based on the number of walks of a given length $\ell$ starting from that node. This measure captures higher-order connectivity patterns within the graph, going beyond direct neighbors to account for walks that traverse intermediate nodes. In this work, we consider the Walk Count centrality for $\ell=2$, as it effectively captures higher-order structures and provides valuable insights into the organization of complex systems \cite{benson2016higher}.

 \smallskip
Using this assumption, we employ node clustering to partition the node set into $C$ clusters, denoted by $\mathcal{P} = \cup_{1\leq c \leq C}  \mathcal{P}_c $. Each cluster $c\in \{1,\ldots,C\}$ is assigned a common exit layer $\ell_c\in \{0,\ldots, L\}$, determined using nodes excluded from the test set. Validation nodes are utilized for this purpose. (i) Centrality scores are computed for all nodes. (ii) Nodes are ranked based on their centrality scores. (iii)  These centrality scores are discretized into $C$ equal-sized buckets to facilitate the clustering process. (iv) The optimal exit layer for each cluster is determined by evaluating the classification accuracy on validation nodes within that cluster, i.e.,
$$ \forall c\in \{1,\ldots,C\}, \quad \ell_c = \argmax_{\ell \in \{0,\ldots, L\}} \text{Acc} \left \{ p^{(\ell)}_v \in \mathcal{V}_{val} \cap \mathcal{P}_c\right \}.$$

\paragraph{Ablation Study.} The choice of using validation nodes rather than training nodes for learning the
layer selection policy stems for the high prediction accuracy on training nodes across all layers. This high accuracy leads to a significant distribution shift between the predictions for train nodes and those for test nodes. Conversely, validation nodes, which were not utilized during the training of the ADMP-GNN, present a more suitable option for learning the policy due to their unbiased predictions.  For the choice of the cluster-based layer selection policy, we could use a variety of deep learning mechanisms to predict the best exit layer for each node. This included generalizing the policy by training neural networks to identify layers that accurately predict node outcomes or by framing the problem as an optimal stopping problem following the framework of \cite{hure2021deep}. However, these approaches require learning the policy on a set of nodes larger than the test set, which is impractical for node classification tasks where the training and validation node sets available for policy learning are typically smaller than the test set.

\section{Experimental Evaluation}
\begin{table*}[t]
\centering
\caption{Classification accuracy ($\pm$ standard deviation) on different node classification datasets for the baselines based on the GCN backbone. The higher the accuracy (in \%), the better the model. Highlighted are the \textbf{first}, \underline{second} best results. OOM means \textit{Out of memory}.}
\resizebox{1.6\columnwidth}{!}{%
\begin{tabular}{llllllll}
\toprule
Model & Cora & CiteSeer & CS & PubMed & Genuis &  Ogbn-arxiv\\ \midrule

 JKNET-CAT  & $79.52 {\scriptstyle \pm 1.16} ~[2]$ & $69.69 {\scriptstyle \pm 0.05~[1]} $ & $91.23 {\scriptstyle \pm 1.26~[1]} $ & $77.63 {\scriptstyle \pm 0.59} ~[2]$ & $\mathbf{81.46 {\scriptstyle \pm 0.10} ~[2]}$ & $68.54 {\scriptstyle \pm 0.57} ~[5]$ \\

JKNET-MAX  & $75.67 {\scriptstyle \pm 0.18} ~[1]$ & $70.12 {\scriptstyle \pm 0.04} ~[1]$ & $90.55 {\scriptstyle \pm 0.74} ~[1]$ & $75.10 {\scriptstyle \pm 2.41} ~[2]$ & $80.13 {\scriptstyle \pm 0.29} ~[1]$ & $56.66 {\scriptstyle \pm 0.31} ~[5]$ \\

JKNET-LSTM  & $78.95 {\scriptstyle \pm 0.62} ~[0]$ & $65.83 {\scriptstyle \pm 1.27} ~[0]$ & $90.17 {\scriptstyle \pm 1.41} ~[2]$ & $77.73 {\scriptstyle \pm 0.67} ~[0]$ & OOM & OOM \\

Residuals - GCNII  & $76.84 {\scriptstyle \pm 0.20} ~[1]$ & $69.72 {\scriptstyle \pm 0.06} ~[1]$ & $90.84 {\scriptstyle \pm 1.35} ~[1]$ & $77.82 {\scriptstyle \pm 0.44} ~[2]$ & \underline{$81.36 {\scriptstyle \pm 1.13} ~[2]$} & $61.48 {\scriptstyle \pm 3.10} ~[4]$ \\

AdaGCN  & $75.08 {\scriptstyle \pm 0.27}~[1]$  & $69.58 {\scriptstyle \pm 0.19}~[1]$  & $89.62 {\scriptstyle \pm 0.51}~[0] $ & $76.40 {\scriptstyle \pm 0.12}~[5]$ & $79.85 {\scriptstyle \pm 0.00}~[0]$ & $22.06 {\scriptstyle \pm 1.67}~[5]$ \\

GPR-GNN  & $79.91 {\scriptstyle \pm 0.43} ~[2]$ & $69.21 {\scriptstyle \pm 0.81} ~[2]$ & $91.42 {\scriptstyle \pm 1.12} ~[2]$ & $79.00 {\scriptstyle \pm 0.39} ~[2]$ & $81.04 {\scriptstyle \pm 0.41} ~[2]$ & $68.03 {\scriptstyle \pm 0.23} ~[3]$ \\ \midrule

GCN & $80.78 {\scriptstyle \pm 0.72} ~[2]$ & $71.25 {\scriptstyle \pm 0.72} ~[2]$ & $\mathbf{92.20 {\scriptstyle \pm 0.00} ~[1]}$ & $\mathbf{79.32 {\scriptstyle \pm 0.41} ~[2]}$ & $80.76 {\scriptstyle \pm 1.05} ~[0]$ & $64.37 {\scriptstyle \pm 0.43} ~[2]$ \\
 ADMP-GCN & \underline{$81.22 {\scriptstyle \pm 0.36} ~[5]$} & $\mathbf{71.33 {\scriptstyle \pm 0.40} ~[2]}$ & \underline{$91.49 {\scriptstyle \pm 0.66} ~[2]$} & \underline{$79.02 {\scriptstyle \pm 0.21} ~[2]$} & $80.17 {\scriptstyle \pm 0.79} ~[0]$ & $69.29 {\scriptstyle \pm 0.74} ~[4]$ \\ \midrule

ADMP-GCN w/ \textit{Degree} & $81.03 {\scriptstyle \pm 0.53}$ & $71.10 {\scriptstyle \pm 0.50}$ & $91.26 {\scriptstyle \pm 0.59}$ & $78.71 {\scriptstyle \pm 0.39}$ & $80.73 {\scriptstyle \pm 1.00}$ & \underline{$69.59 {\scriptstyle \pm 0.28}$} \\

ADMP-GCN w/ \textit{$k$-core} & $\mathbf{81.19 {\scriptstyle \pm 0.40}}$ & \underline{$71.27 {\scriptstyle \pm 0.53}$} & $91.29 {\scriptstyle \pm 0.68}$ & $78.73 {\scriptstyle \pm 0.41}$ & $80.73 {\scriptstyle \pm 1.00}$ & $69.55 {\scriptstyle \pm 0.33}$ \\

ADMP-GCN w/ \textit{Walk Count} & $81.14 {\scriptstyle \pm 0.41}$ & $71.14 {\scriptstyle \pm 0.50}$ & $91.19 {\scriptstyle \pm 0.64}$ & $78.64 {\scriptstyle \pm 0.60}$ & $80.68 {\scriptstyle \pm 0.97}$ & $69.55 {\scriptstyle \pm 0.34}$ \\

ADMP-GCN w/ \textit{PageRank} & $81.05 {\scriptstyle \pm 0.46}$ & $71.00 {\scriptstyle \pm 0.29}$ & $91.09 {\scriptstyle \pm 0.99}$ & $78.69 {\scriptstyle \pm 0.56}$ & $81.12 {\scriptstyle \pm 1.41}$ & $\mathbf{69.60 {\scriptstyle \pm 0.29}}$ \\
 \bottomrule
\end{tabular}
}

\label{tab:results_with_baselines}
\end{table*}

\begin{table*}[h]
\centering
\caption{Classification accuracy ($\pm$ standard deviation) on different node classification datasets for the baselines based on the GIN backbone. The higher the accuracy (in \%), the better the model. Highlighted are the \textbf{first}, \underline{second} best results. OOM means \textit{Out of memory}.}
\resizebox{1.6\columnwidth}{!}{%
\begin{tabular}{llllllll}
\toprule
Model & Cora & CiteSeer & CS & PubMed & Genuis &  Ogbn-arxiv\\ \midrule

 JKNET-CAT  & $77.94 {\scriptstyle \pm 0.67} ~[2]$ & $64.82 {\scriptstyle \pm 0.04} ~[1]$ & $89.26 {\scriptstyle \pm 1.18} ~[1]$ & $75.89 {\scriptstyle \pm 2.50} ~[2]$ & OOM & $60.23 {\scriptstyle \pm 0.37} ~[1]$ \\

  JKNET-MAX  & $77.47 {\scriptstyle \pm 0.81} ~[1]$ & $64.96 {\scriptstyle \pm 2.08} ~[2]$ & $87.40 {\scriptstyle \pm 2.11} ~[0]$ & $76.21 {\scriptstyle \pm 1.73} ~[0]$ & OOM & $51.84 {\scriptstyle \pm 4.01} ~[0]$ \\

JKNET-LSTM & $77.39 {\scriptstyle \pm 1.39} ~[2]$ & $64.52 {\scriptstyle \pm 2.02} ~[1]$ & $87.27 {\scriptstyle \pm 3.66} ~[3]$ & $75.90 {\scriptstyle \pm 1.63} ~[0]$ & OOM & OOM \\

GPR-GIN & $76.83 {\scriptstyle \pm 1.22} ~[2]$ & \underline{$66.43 {\scriptstyle \pm 1.15} ~[2]$} & $88.15 {\scriptstyle \pm 1.53} ~[5]$ & $\mathbf{77.27 {\scriptstyle \pm 0.87} ~[2]}$ & $80.82 {\scriptstyle \pm 0.42} ~[3]$ & $\mathbf{63.05 {\scriptstyle \pm 0.44} ~[2]}$ \\  \midrule

GIN & $77.73 {\scriptstyle \pm 0.99} ~[2]$ & $65.23 {\scriptstyle \pm 1.45} ~[2]$ & $90.29 {\scriptstyle \pm 0.99} ~[1]$ & $76.05 {\scriptstyle \pm 1.14} ~[2]$ & $80.78 {\scriptstyle \pm 1.03} ~[0]$ & $60.70 {\scriptstyle \pm 0.15} ~[1]$ \\

 ADMP-GIN & $78.07 {\scriptstyle \pm 0.68} ~[2]$ & $65.41 {\scriptstyle \pm 1.91} ~[2]$ & $\underline{90.82 {\scriptstyle \pm 1.15} ~[1]}$ & $76.46 {\scriptstyle \pm 1.04} ~[4]$ & $80.47 {\scriptstyle \pm 0.91} ~[0]$ & $60.85 {\scriptstyle \pm 0.01} ~[1]$ \\ \midrule

 ADMP-GIN w/ \textit{Degree} & \underline{$78.12 {\scriptstyle \pm 0.70}$} & $66.82 {\scriptstyle \pm 0.89}$ & $90.70 {\scriptstyle \pm 0.79}$ & $76.07 {\scriptstyle \pm 1.27}$ & \underline{$81.68 {\scriptstyle \pm 0.70}$} & $60.85 {\scriptstyle \pm 0.01}$ \\

ADMP-GIN w/ \textit{$k$-core} & $78.10 {\scriptstyle \pm 0.64}$ & $66.36 {\scriptstyle \pm 1.03}$ & $\mathbf{90.85 {\scriptstyle \pm 0.93}}$ & $75.93 {\scriptstyle \pm 1.13}$ & $81.48 {\scriptstyle \pm 0.64}$ & \underline{$60.85 {\scriptstyle \pm 0.01}$} \\

ADMP-GIN w/ \textit{Walk Count} & $\mathbf{78.19 {\scriptstyle \pm 0.68}}$ & $65.79 {\scriptstyle \pm 0.81}$ & $90.77 {\scriptstyle \pm 0.91}$ & \underline{$76.72 {\scriptstyle \pm 0.76}$} & $81.21 {\scriptstyle \pm 0.58}$ & $60.85 {\scriptstyle \pm 0.01}$ \\

ADMP-GIN w/ \textit{PageRank} & $77.78 {\scriptstyle \pm 0.83}$ & $\mathbf{67.08 {\scriptstyle \pm 0.51}}$ & $90.72 {\scriptstyle \pm 0.91}$ & $76.63 {\scriptstyle \pm 0.87}$ & $\mathbf{81.89 {\scriptstyle \pm 1.04}}$ & $60.85 {\scriptstyle \pm 0.01}$ \\

 \bottomrule

\end{tabular}
}
\label{tab:results_with_baselines_GIN}
\end{table*}

We now discuss the experimental setup and results in turn. 
\subsection{Experimental Setup}
\paragraph{Datasets.} We use thirteen widely used datasets in the GNN literature. We particularly used the citation networks Cora, CiteSeer, and PubMed \cite{dataset_node_classification}, the co-authorship networks CS \cite{cs_data}, the citation network between Computer Science arXiv papers Ogbn-arxiv  \cite{hu2020open}, the Amazon Computers and Amazon Photo networks \cite{cs_data}, the non-homophilous dataset genius \cite{lim2021expertise}, and the disassortative datasets Chameleon, Squirrel \cite{rozemberczki2021multi}, 
 and Cornell, Texas, Wisconsin from the WebKB dataset \cite{lim2021large}. More details and statistics about the used datasets can be found in Appendix \ref{app:data_stats}. For the Cora, CiteSeer, and Pubmed datasets, we used the provided train/validation/test splits. For the remaining datasets, we followed the framework in \cite{lim2021large,rozemberczki2021multi}.

\paragraph{Baselines.} We compare our approach with architectures that combine all the hidden representations of nodes to form a final node representation used for prediction. For each baseline model, we vary the number of layers from 0 to 5, and we report in Table \ref{tab:results_with_baselines} the performance of the best number of layers with respect to the test set. (i) This includes \textit{Jumping knowledge}, which combines the nodes representation of all layers using an aggregation layer, e.g., MaxPooling (JKMaxPool), Concatenation (JK-Concat), or LSTM-attention (JK-LSTM) \cite{pmlr-v80-xu18c}. (ii) Residuals-GCNII uses an initial residual connection and an identity mapping for each layer.  The initial residual connection ensures that the final representation of each node retains at least a fraction of $\alpha$ from the input layer \cite{DBLP:journals/corr/abs-2007-02133}. (iii)  GPR-GCN combines adaptive generalized PageRank (GPR) scheme with GNNs \cite{chien2020adaptive}. 
(iv)  Ada-GCN, which proposes an RNN-like deep GNN architecture by incorporating AdaBoost to combine the layers \cite{sun2019adagcn}.  In contrast to the baselines, which aggregate the layers and determine the best performance by performing a grid search over the number of layers $L$, we fix the maximum number of layers to $L=5$ for ADMP-GNN. This comparison gives a strong advantage to the baselines, as their results are optimized for each dataset, while ADMP-GNN's performance is evaluated under a fixed and consistent setup.

\paragraph{Implementation Details.} We train all the models using the Adam optimizer \cite{kingma_adam} and the same hyperparameters. The GNN hyperparameters in each dataset were optimized using a grid search on the classical GCN; we detail the values of these hyperparameters in Table \ref{tab:gcn_tuned_hyperparams} of Appendix \ref{app:hyper}. To account for the impact of random initialization, each experiment was repeated 10 times, and the mean and standard deviation of the results were reported. The experiments have been run on both an NVIDIA A100 GPU.

\subsection{Experimental Results}

Through extensive experiments on multiple datasets, we can better understand the scenarios in which ADMP-GNN proves to be effective. As observed in Tables \ref{tab:results_with_baselines}, \ref{tab:results_with_baselines_GIN}, a comparison between ADMP-GCN and ADMP-GIN against their respective baselines, GCN and GIN, demonstrates consistently higher accuracy for most datasets. Regarding the centrality-based layer selection policy, it becomes clear that this policy is particularly efficient when graphs exhibit a wide range of local density and centrality among nodes. Most importantly, no impactful drop in accuracy was observed with ADMP-GNN. It is important to note that the suitability of a given centrality metric can vary significantly across datasets. For instance, while PageRank is widely used, its distribution often forms a peak near zero due to the normalization constraint where the sum of PageRank scores across nodes equals 1, making it challenging to form meaningful clusters. In contrast, other centrality metrics like $k$-core and Walk Count are more adaptable in datasets where clear clustering patterns emerge. For example, in datasets like Cora, ogbn-arxiv, and Photo, the $k$-core centrality effectively highlights clusters, enabling an adaptive depth policy. For datasets like Texas and Wisconsin, Walk Count effectively captures the structural diversity necessary for cluster-based layer selection policy. These observations emphasize the importance of selecting appropriate centrality metrics tailored to the specific properties of a dataset to maximize the effectiveness of ADMP-GNN.

\section{Conclusion}
In this work, we propose ADMP-GNN, a novel adaption of message passing neural networks that enables to make predictions for each node at every layer. Additionally, we propose a sequential training approach designed to achieve performance comparable to training multiple GNNs independently in a single-task setting.  Our empirical analysis highlights the importance of node-specific depth in GNNs to effectively capture the unique characteristics and computational requirements of each node. Determining the optimal number of message-passing layers for each node is a challenging task, influenced by factors such as the complexity and connectivity of graph structures and the variability introduced by node features. Through our experiments, we identified that node centrality can serve as a useful indicator for determining the optimal layer for each node. To this end, we heuristically learn a layer selection policy using a set of validation nodes, which is then generalized to test nodes. Extensive experiments across multiple datasets, particularly those characterized by a diversity in local structural properties, demonstrate that ADMP-GNN significantly enhances the prediction accuracy of GNNs, offering an effective solution to address the challenges of layer selection and node-specific learning.

\section*{Acknowledgment} F.M. acknowledges the support of the Innov4-ePiK project managed by the French National Research Agency under the 4th PIA, integrated into France2030 (ANR-23-RHUS-0002). Y.A. and M.V. are supported by the French National Research Agency (ANR) via the AML-HELAS (ANR-19-CHIA-0020) project.  Y.A. and J.L. are supported by the French National Research Agency (ANR) via the ``GraspGNNs'' JCJC grant (ANR-24-CE23-3888). This work was granted access to the HPC resources of IDRIS under the allocation
“2024-AD010613410R2” made by GENCI.

\appendix

\section{Node-Specific Depth Analysis in Graph Neural Networks}\label{app:motivation}
We generate synthetic graphs of size $N=5,000$. We select nodes belonging to sparse or dense regions in the original graph based on their core number. We consider only nodes with labels that are sufficiently present in both dense and sparse region. Last, we randomly select nodes, all by keeping the label distribution similar in both sparse and dense subgraphs. Yellow points indicate edges, while purple points represent non-edges. Notably, there are variations in edge density across different blocks: the first block is characterized by extreme sparsity, whereas the second block exhibits a much denser structure.

\begin{figure}[ht]
    \centering
    \includegraphics[width=\linewidth]{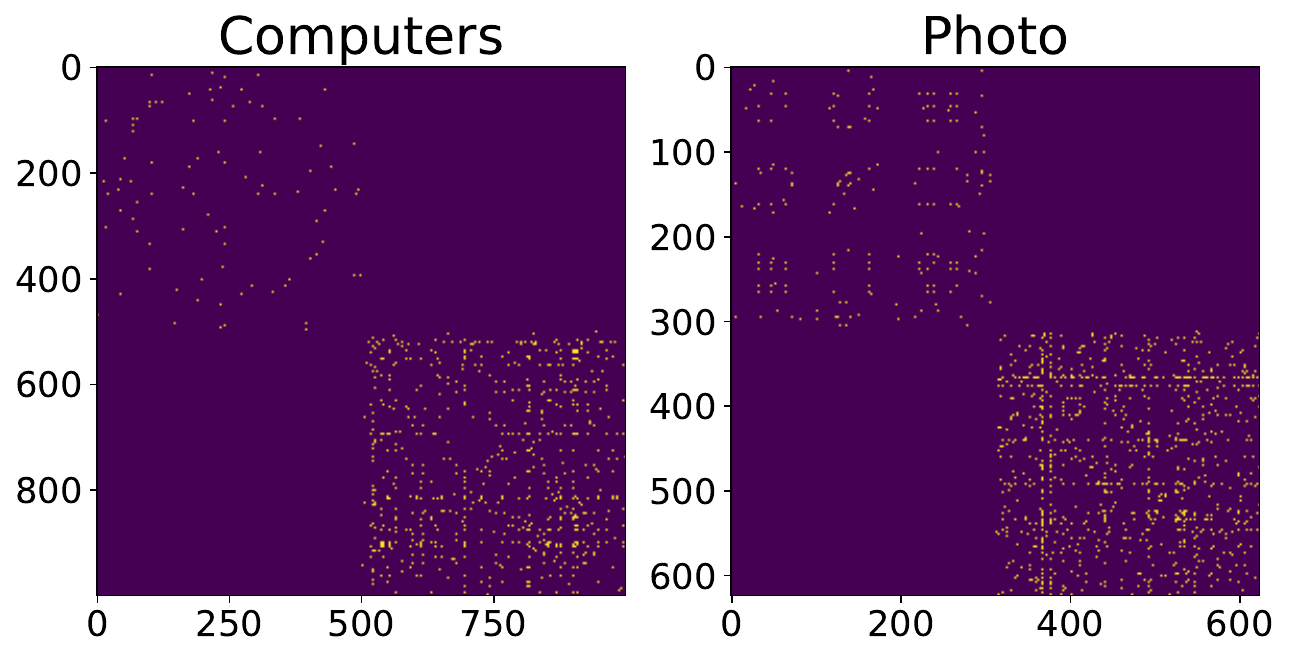}
    \caption{The adjacency matrix of the synthetic graphs extracted from the real graphs Computers and Photo.}
    \label{fig:motivation_hybid}
\end{figure}

\section{Time Complexity} \label{app:time_comp}
Understanding the time complexity of the ADMP-GCN model is crucial for evaluating its practical efficiency and scalability. Table \ref{tab:time_complexity} reports the average training time, measured in seconds, for two distinct settings: ALM  and ST.

\begin{table}[h]
\centering
\caption{The average time needed for each training setting for different datasets. }
\resizebox{\columnwidth}{!}{%
\begin{tabular}{llllllll}
\toprule
Model & Cora & squirel & chamelon & Computers  & Photo & Ogbn-arxiv  \\ \midrule
ADMP-GCN ALM & 14 & 36 & 15 & 51 & 26 & 266 \\
ADMP-GCN ST &32 & 88 & 40 & 175 & 87 & 882  \\ 
\bottomrule

\end{tabular}
}
\label{tab:time_complexity}
\end{table}

\section{Dataset Statistics}\label{app:data_stats}
Characteristics and information about the datasets utilized in the node classification part of the study are presented in Table \ref{tab:data_statistics}, which provides statistics about each dataset, and  highlights the variety in their features, size, and edge homophily,

\begin{table}[h]

\caption{Statistics of the node classification datasets used in our experiments.}
\label{tab:data_statistics}
\begin{center}
\begin{small}
\resizebox{\columnwidth}{!}{%
\begin{tabular}{lrrrrr}
\toprule
Dataset & \#Features & \#Nodes & \#Edges & \#Classes & Edge Homophily \\
\midrule
Cora    & 1,433 & 2,708   & 5,208    & 7 & 0.809 \\
CiteSeer   & 3,703 & 3,327 & 4,552 & 6 & 0.735\\
PubMed    & 500 & 19,717 & 44,338 & 3 & 0.802\\
CS    & 6,805 & 18,333 & 81,894 & 15 & 0.808\\

Genuis  & 12    &   421,961  & 984,979   &  5   &   0.618  \\
Ogbn-arxiv  &   128  &   169,343  &   2,315,598  &   40  &   0.654  \\

\hline
\end{tabular}
}
\end{small}
\end{center}
\end{table}

\section{Hyperparameter Configurations}\label{app:hyper}
For a more balanced comparison, however, we use the same training procedure for all the models. The hyperparameters in each dataset where optimized using a grid search on the classical GCN over the following search space:
\begin{itemize}
    \item Hidden size: $[16, 32, 64, 128, 256, 512],$,
    \item Learning rate: $[0.1, 0.01, 0.001],$
    \item Dropout probability: $[16, 32, 64, 128, 256, 512].$
\end{itemize}

The number of layers was fixed to 2. The optimal hyperparameters can be found in Table \ref{tab:gcn_tuned_hyperparams}.

\begin{table}[h]
\caption{Hyperparameters used in our experiments.}
\label{tab:gcn_tuned_hyperparams}
\begin{center}
\begin{small}
\resizebox{0.95\columnwidth}{!}{%
\begin{tabular}{lccc}
\toprule
Dataset & Hidden Size & Learning Rate & Dropout Probability   \\
\midrule
Cora    & 64 & 0.01 & 0.8   \\
CiteSeer    &64 & 0.01 & 0.4   \\
PubMed     &64 & 0.01 & 0.2   \\\
CS     & 512 & 0.01 & 0.4 \\
Genuis   & 512 & 0.01  & 0.2 \\
Ogbn-Arxiv    & 512 & 0.01 & 0.5  \\

\bottomrule
\end{tabular}
}
\end{small}
\end{center}
\end{table}


\section{Supplementary Results of ADMP-GIN}\label{app:GIN}

We replicate the experiments from the main papers described in Section \ref{method}, using the GIN backbone. The results for ADMP-GIN are presented in Tables \ref{tab:multi_task_GIN} and \ref{tab:oracle_GIN}.

\begin{table*}[h]
\caption{Comparison of ADMP-GIN training paradigms ALM and ST. These paradigms are also compared to the single-task training setting to evaluate which approach most closely mimics the classical GIN under single-task training.  The best  multi-task results for each dataset are \textbf{bolded}.}
\resizebox{0.75\textwidth}{!}{%
\begin{tabular}{c|clcccccc}
\toprule
\# Layers & Training Paradigm & Model & Cora & CiteSeer & CS & PubMed & Genius & Ogbn-arxiv \\
\midrule

\multirow{3}{*}{\circled{0}} & \multirow{1}{*}{Single-task} & GIN & $56.40 {\scriptstyle \pm 0.06}$ & $57.14 {\scriptstyle \pm 0.09}$ & $87.17 {\scriptstyle \pm 1.41}$ & $72.49 {\scriptstyle \pm 0.08}$ & $80.78 {\scriptstyle \pm 1.03}$ & $48.87 {\scriptstyle \pm 0.04}$ \\ \cmidrule(lr){2-9}
& \multirow{2}{*}{Multi-task} & ADMP-GIN (ALM) & $\mathbf{58.00 {\scriptstyle \pm 0.00}}$ & $\mathbf{61.50 {\scriptstyle \pm 0.00}}$ & $86.36 {\scriptstyle \pm 0.78}$ & $\mathbf{73.20 {\scriptstyle \pm 0.00}}$ & $79.95 {\scriptstyle \pm 0.10}$ & $36.49 {\scriptstyle \pm 0.19}$ \\
& & ADMP-GIN (ST)  & $56.38 {\scriptstyle \pm 0.04}$ & $57.17 {\scriptstyle \pm 0.08}$ & $\mathbf{87.41 {\scriptstyle \pm 0.95}}$ & $72.47 {\scriptstyle \pm 0.14}$ & $\mathbf{80.47 {\scriptstyle \pm 0.91}}$ & $\mathbf{48.87 {\scriptstyle \pm 0.04}}$ \\ \midrule

\multirow{3}{*}{\circled{1}} & \multirow{1}{*}{Single-task} & GIN & $75.17 {\scriptstyle \pm 0.09}$ & $64.79 {\scriptstyle \pm 0.03}$ & $90.29 {\scriptstyle \pm 0.99}$ & $74.97 {\scriptstyle \pm 0.11}$ & $78.42 {\scriptstyle \pm 4.95}$ & $60.90 {\scriptstyle \pm 0.15}$ \\ \cmidrule(lr){2-9}
& \multirow{2}{*}{Multi-task} & ADMP-GIN (ALM) & $74.50 {\scriptstyle \pm 0.00}$ & $\mathbf{66.50 {\scriptstyle \pm 0.00}}$ & $88.66 {\scriptstyle \pm 1.18}$ & $\mathbf{75.40 {\scriptstyle \pm 0.00}}$ & $72.07 {\scriptstyle \pm 17.44}$ & $59.43 {\scriptstyle \pm 0.73}$ \\
& & ADMP-GIN (ST)  & $\mathbf{75.07 {\scriptstyle \pm 0.06}}$ & $64.80 {\scriptstyle \pm 0.00}$ & $\mathbf{90.82 {\scriptstyle \pm 1.15}}$ & $75.00 {\scriptstyle \pm 0.00}$ & $\mathbf{77.01 {\scriptstyle \pm 12.06}}$ & $\mathbf{60.85 {\scriptstyle \pm 0.01}}$ \\ \midrule

\multirow{3}{*}{\circled{2}} & \multirow{1}{*}{Single-task} & GIN & \textbf{$77.73 {\scriptstyle \pm 0.99}$} & \textbf{$65.23 {\scriptstyle \pm 1.45}$} & $87.93 {\scriptstyle \pm 0.71}$ & $76.05 {\scriptstyle \pm 1.14}$ & $78.89 {\scriptstyle \pm 0.48}$ & $16.23 {\scriptstyle \pm 9.60}$\\ \cmidrule(lr){2-9}
& \multirow{2}{*}{Multi-task} &ADMP-GIN (ALM) & $63.53 {\scriptstyle \pm 4.80}$ & $60.68 {\scriptstyle \pm 2.09}$ & $12.90 {\scriptstyle \pm 8.05}$ & $72.20 {\scriptstyle \pm 4.11}$ & $\mathbf{79.60 {\scriptstyle \pm 0.51}}$ & $\mathbf{47.63 {\scriptstyle \pm 3.20}}$ \\
& & ADMP-GIN (ST)  & $\mathbf{78.07 {\scriptstyle \pm 0.68}}$ & $\mathbf{65.41 {\scriptstyle \pm 1.91}}$ & $\mathbf{87.47 {\scriptstyle \pm 2.18}}$ & $\mathbf{76.04 {\scriptstyle \pm 0.88}}$ & $72.99 {\scriptstyle \pm 13.48}$ & $10.13 {\scriptstyle \pm 8.00}$ \\ \midrule

\multirow{3}{*}{\circled{3}} & \multirow{1}{*}{Single-task} & GIN & $74.40 {\scriptstyle \pm 1.14}$ & $60.81 {\scriptstyle \pm 2.20}$ & $82.13 {\scriptstyle \pm 2.24}$ & $74.97 {\scriptstyle \pm 1.69}$ & $52.22 {\scriptstyle \pm 28.47}$ & $6.00 {\scriptstyle \pm 0.17}$ \\ \cmidrule(lr){2-9}
& \multirow{2}{*}{Multi-task} &ADMP-GIN (ALM) & $66.99 {\scriptstyle \pm 2.85}$ & $59.57 {\scriptstyle \pm 2.57}$ & $17.69 {\scriptstyle \pm 12.02}$ & $74.15 {\scriptstyle \pm 2.00}$ & $68.00 {\scriptstyle \pm 23.98}$ & $\mathbf{27.51 {\scriptstyle \pm 16.25}}$ \\
& & ADMP-GIN (ST)  & $\mathbf{76.28 {\scriptstyle \pm 1.11}}$ & $\mathbf{65.13 {\scriptstyle \pm 1.00}}$ & $\mathbf{83.60 {\scriptstyle \pm 3.72}}$ & $\mathbf{76.21 {\scriptstyle \pm 1.75}}$ & $\mathbf{80.04 {\scriptstyle \pm 0.09}}$ & $13.74 {\scriptstyle \pm 9.66}$ \\ \midrule

\multirow{3}{*}{\circled{4}} & \multirow{1}{*}{Single-task} & GIN & $67.68 {\scriptstyle \pm 4.07}$ & $57.54 {\scriptstyle \pm 2.92}$ & $47.37 {\scriptstyle \pm 17.10}$ & $73.62 {\scriptstyle \pm 1.63}$ & \textbf{$80.05 {\scriptstyle \pm 0.10}$} & $6.07 {\scriptstyle \pm 0.21}$ \\ \cmidrule(lr){2-9}
& \multirow{2}{*}{Multi-task} & ADMP-GIN (ALM) & $68.78 {\scriptstyle \pm 4.30}$ & $60.12 {\scriptstyle \pm 2.36}$ & $21.07 {\scriptstyle \pm 14.15}$ & $74.20 {\scriptstyle \pm 2.57}$ & $79.36 {\scriptstyle \pm 1.18}$ & $15.23 {\scriptstyle \pm 11.85}$ \\
& & ADMP-GIN (ST)  & $\mathbf{74.94 {\scriptstyle \pm 1.58}}$ & $\mathbf{65.21 {\scriptstyle \pm 1.54}}$ & $\mathbf{81.33 {\scriptstyle \pm 2.15}}$ & $\mathbf{76.46 {\scriptstyle \pm 1.04}}$ & $\mathbf{80.04 {\scriptstyle \pm 0.09}}$ & $\mathbf{16.02 {\scriptstyle \pm 10.78}}$ \\ \midrule

\multirow{3}{*}{\circled{5}} & \multirow{1}{*}{Single-task} & GIN & $32.48 {\scriptstyle \pm 10.47}$ & $54.76 {\scriptstyle \pm 2.32}$ & $18.56 {\scriptstyle \pm 11.32}$ & $67.98 {\scriptstyle \pm 5.80}$ & $80.05 {\scriptstyle \pm 0.10}$ & $6.19 {\scriptstyle \pm 0.39}$ \\ \cmidrule(lr){2-9}
& \multirow{2}{*}{Multi-task} &ADMP-GIN (ALM) & $67.46 {\scriptstyle \pm 4.34}$ & $59.74 {\scriptstyle \pm 1.95}$ & $29.16 {\scriptstyle \pm 13.84}$ & $73.94 {\scriptstyle \pm 2.42}$ & $\mathbf{79.99 {\scriptstyle \pm 0.10}}$ & $14.61 {\scriptstyle \pm 12.11}$ \\
& & ADMP-GIN (ST)  & $\mathbf{71.34 {\scriptstyle \pm 2.09}}$ & $\mathbf{63.89 {\scriptstyle \pm 1.39}}$ & $\mathbf{79.10 {\scriptstyle \pm 1.84}}$ & $\mathbf{75.75 {\scriptstyle \pm 1.09}}$ & $79.57 {\scriptstyle \pm 1.40}$ & $\mathbf{15.72 {\scriptstyle \pm 12.13}}$ \\ 
\bottomrule

\end{tabular}
}
\label{tab:multi_task_GIN}
\end{table*}

\begin{table*}[h]
\centering
\caption{Comparison of highest accuracy ($\pm$ standard deviation) for GIN and ADMP-GIN ST, with the layer achieving the best accuracy indicated in brackets. The final row shows the \textit{Oracle Accuracy} for ADMP-GIN ST. Best results for each dataset are \textbf{bolded}.}
\resizebox{0.8\textwidth}{!}{%
\begin{tabular}{lllllll}
\toprule
 Model & Cora & CiteSeer & CS & PubMed & Genius & Ogbn-arxiv \\  \midrule

GIN & $77.73 {\scriptstyle \pm 0.99} ~[2]$ & $65.23 {\scriptstyle \pm 1.45} ~[2]$ & $90.29 {\scriptstyle \pm 0.99} ~[1]$ & $76.05 {\scriptstyle \pm 1.14} ~[2]$ & $80.78 {\scriptstyle \pm 1.03} ~[0]$ & $60.90 {\scriptstyle \pm 0.15} ~[1]$ \\

ADMP-GIN (ST) & $78.07 {\scriptstyle \pm 0.68} ~[2]$ & $65.41 {\scriptstyle \pm 1.91} ~[2]$ & $90.82 {\scriptstyle \pm 1.15} ~[1]$ & $76.46 {\scriptstyle \pm 1.04} ~[4]$ & $80.47 {\scriptstyle \pm 0.91} ~[0]$ & $60.85 {\scriptstyle \pm 0.01} ~[1]$ \\

ADMP-GIN ST - Oracle & $\mathbf{90.76} {\scriptstyle \pm \mathbf{0.27}}$ & $\mathbf{81.87} {\scriptstyle \pm \mathbf{0.63}}$ & $\mathbf{97.64} {\scriptstyle \pm \mathbf{0.18}}$ & $\mathbf{92.73} {\scriptstyle \pm \mathbf{0.54}}$ & $\mathbf{92.07} {\scriptstyle \pm \mathbf{6.13}}$ & $\mathbf{71.23} {\scriptstyle \pm \mathbf{2.69}}$ \\   \bottomrule

\end{tabular}
}
\label{tab:oracle_GIN}
\end{table*}

\section{GenAI Usage Disclosure}
No generative AI tools were used in the preparation of this research, including in the code, data, analysis, or writing of this paper.

\newpage
\bibliographystyle{ACM-Reference-Format}
\bibliography{bibfile}


\begin{thebibliography}{49}


\ifx \showCODEN    \undefined \def \showCODEN     #1{\unskip}     \fi
\ifx \showISBNx    \undefined \def \showISBNx     #1{\unskip}     \fi
\ifx \showISBNxiii \undefined \def \showISBNxiii  #1{\unskip}     \fi
\ifx \showISSN     \undefined \def \showISSN      #1{\unskip}     \fi
\ifx \showLCCN     \undefined \def \showLCCN      #1{\unskip}     \fi
\ifx \shownote     \undefined \def \shownote      #1{#1}          \fi
\ifx \showarticletitle \undefined \def \showarticletitle #1{#1}   \fi
\ifx \showURL      \undefined \def \showURL       {\relax}        \fi
\providecommand\bibfield[2]{#2}
\providecommand\bibinfo[2]{#2}
\providecommand\natexlab[1]{#1}
\providecommand\showeprint[2][]{arXiv:#2}

\bibitem[Benson et~al\mbox{.}(2016)]%
        {benson2016higher}
\bibfield{author}{\bibinfo{person}{Austin~R Benson}, \bibinfo{person}{David~F Gleich}, {and} \bibinfo{person}{Jure Leskovec}.} \bibinfo{year}{2016}\natexlab{}.
\newblock \showarticletitle{Higher-order organization of complex networks}.
\newblock \bibinfo{journal}{\emph{Science}} \bibinfo{volume}{353}, \bibinfo{number}{6295} (\bibinfo{year}{2016}), \bibinfo{pages}{163--166}.
\newblock


\bibitem[Bolukbasi et~al\mbox{.}(2017)]%
        {bolukbasi2017adaptive}
\bibfield{author}{\bibinfo{person}{Tolga Bolukbasi}, \bibinfo{person}{Joseph Wang}, \bibinfo{person}{Ofer Dekel}, {and} \bibinfo{person}{Venkatesh Saligrama}.} \bibinfo{year}{2017}\natexlab{}.
\newblock \showarticletitle{Adaptive neural networks for efficient inference}. In \bibinfo{booktitle}{\emph{International Conference on Machine Learning}}. PMLR, \bibinfo{pages}{527--536}.
\newblock


\bibitem[Bornholdt and Schuster(2001)]%
        {bornholdt2001handbook}
\bibfield{author}{\bibinfo{person}{Stefan Bornholdt} {and} \bibinfo{person}{Heinz~Georg Schuster}.} \bibinfo{year}{2001}\natexlab{}.
\newblock \showarticletitle{Handbook of graphs and networks}.
\newblock \bibinfo{journal}{\emph{From Genome to the Internet, Willey-VCH (2003 Weinheim)}} (\bibinfo{year}{2001}).
\newblock


\bibitem[Brin and Page(1998)]%
        {brin1998anatomy}
\bibfield{author}{\bibinfo{person}{Sergey Brin} {and} \bibinfo{person}{Lawrence Page}.} \bibinfo{year}{1998}\natexlab{}.
\newblock \showarticletitle{The anatomy of a large-scale hypertextual web search engine}.
\newblock \bibinfo{journal}{\emph{Computer networks and ISDN systems}} \bibinfo{volume}{30}, \bibinfo{number}{1-7} (\bibinfo{year}{1998}), \bibinfo{pages}{107--117}.
\newblock


\bibitem[Cao et~al\mbox{.}(2020)]%
        {cao2020spectral}
\bibfield{author}{\bibinfo{person}{Defu Cao}, \bibinfo{person}{Yujing Wang}, \bibinfo{person}{Juanyong Duan}, \bibinfo{person}{Ce Zhang}, \bibinfo{person}{Xia Zhu}, \bibinfo{person}{Congrui Huang}, \bibinfo{person}{Yunhai Tong}, \bibinfo{person}{Bixiong Xu}, \bibinfo{person}{Jing Bai}, \bibinfo{person}{Jie Tong}, {et~al\mbox{.}}} \bibinfo{year}{2020}\natexlab{}.
\newblock \showarticletitle{Spectral temporal graph neural network for multivariate time-series forecasting}.
\newblock \bibinfo{journal}{\emph{Advances in neural information processing systems}}  \bibinfo{volume}{33} (\bibinfo{year}{2020}), \bibinfo{pages}{17766--17778}.
\newblock


\bibitem[Castro-Correa et~al\mbox{.}(2024)]%
        {castro-correa-tnnls24}
\bibfield{author}{\bibinfo{person}{Jhon~A. Castro-Correa}, \bibinfo{person}{Jhony~H. Giraldo}, \bibinfo{person}{Mohsen Badiey}, {and} \bibinfo{person}{Fragkiskos~D. Malliaros}.} \bibinfo{year}{2024}\natexlab{}.
\newblock \showarticletitle{Gegenbauer Graph Neural Networks for Time-Varying Signal Reconstruction}.
\newblock \bibinfo{journal}{\emph{IEEE Transactions on Neural Networks and Learning Systems}} \bibinfo{volume}{35}, \bibinfo{number}{9} (\bibinfo{year}{2024}), \bibinfo{pages}{11734--11745}.
\newblock


\bibitem[Chen et~al\mbox{.}(2020)]%
        {DBLP:journals/corr/abs-2007-02133}
\bibfield{author}{\bibinfo{person}{Ming Chen}, \bibinfo{person}{Zhewei Wei}, \bibinfo{person}{Zengfeng Huang}, \bibinfo{person}{Bolin Ding}, {and} \bibinfo{person}{Yaliang Li}.} \bibinfo{year}{2020}\natexlab{}.
\newblock \showarticletitle{Simple and deep graph convolutional networks}. In \bibinfo{booktitle}{\emph{International conference on machine learning}}. PMLR, \bibinfo{pages}{1725--1735}.
\newblock


\bibitem[Chien et~al\mbox{.}(2020)]%
        {chien2020adaptive}
\bibfield{author}{\bibinfo{person}{Eli Chien}, \bibinfo{person}{Jianhao Peng}, \bibinfo{person}{Pan Li}, {and} \bibinfo{person}{Olgica Milenkovic}.} \bibinfo{year}{2020}\natexlab{}.
\newblock \showarticletitle{Adaptive universal generalized pagerank graph neural network}.
\newblock \bibinfo{journal}{\emph{arXiv preprint arXiv:2006.07988}} (\bibinfo{year}{2020}).
\newblock


\bibitem[Corso et~al\mbox{.}(2023)]%
        {corso2022diffdock}
\bibfield{author}{\bibinfo{person}{Gabriele Corso}, \bibinfo{person}{Bowen Jing}, \bibinfo{person}{Regina Barzilay}, \bibinfo{person}{Tommi Jaakkola}, {et~al\mbox{.}}} \bibinfo{year}{2023}\natexlab{}.
\newblock \showarticletitle{DiffDock: Diffusion Steps, Twists, and Turns for Molecular Docking}. In \bibinfo{booktitle}{\emph{International Conference on Learning Representations (ICLR 2023)}}.
\newblock


\bibitem[Duval et~al\mbox{.}(2023)]%
        {pmlr-v202-duval23a}
\bibfield{author}{\bibinfo{person}{Alexandre Duval}, \bibinfo{person}{Victor Schmidt}, \bibinfo{person}{Alex Hern\'{a}ndez-Garc\'{\i}a}, \bibinfo{person}{Santiago Miret}, \bibinfo{person}{Fragkiskos~D. Malliaros}, \bibinfo{person}{Yoshua Bengio}, {and} \bibinfo{person}{David Rolnick}.} \bibinfo{year}{2023}\natexlab{}.
\newblock \showarticletitle{{FAEN}et: Frame Averaging Equivariant {GNN} for Materials Modeling}. In \bibinfo{booktitle}{\emph{Proceedings of the 40th International Conference on Machine Learning}}. \bibinfo{publisher}{PMLR}, \bibinfo{pages}{9013--9033}.
\newblock


\bibitem[Elbayad et~al\mbox{.}(2020)]%
        {Elbayad2020Depth-Adaptive}
\bibfield{author}{\bibinfo{person}{Maha Elbayad}, \bibinfo{person}{Jiatao Gu}, \bibinfo{person}{Edouard Grave}, {and} \bibinfo{person}{Michael Auli}.} \bibinfo{year}{2020}\natexlab{}.
\newblock \showarticletitle{Depth-Adaptive Transformer}. In \bibinfo{booktitle}{\emph{International Conference on Learning Representations}}.
\newblock
\urldef\tempurl%
\url{https://openreview.net/forum?id=SJg7KhVKPH}
\showURL{%
\tempurl}


\bibitem[Eliasof et~al\mbox{.}(2024)]%
        {eliasof2024granola}
\bibfield{author}{\bibinfo{person}{Moshe Eliasof}, \bibinfo{person}{Beatrice Bevilacqua}, \bibinfo{person}{Carola-Bibiane Sch{\"o}nlieb}, {and} \bibinfo{person}{Haggai Maron}.} \bibinfo{year}{2024}\natexlab{}.
\newblock \showarticletitle{GRANOLA: Adaptive Normalization for Graph Neural Networks}.
\newblock \bibinfo{journal}{\emph{arXiv preprint arXiv:2404.13344}} (\bibinfo{year}{2024}).
\newblock


\bibitem[Errica et~al\mbox{.}(2023)]%
        {errica2023adaptive}
\bibfield{author}{\bibinfo{person}{Federico Errica}, \bibinfo{person}{Henrik Christiansen}, \bibinfo{person}{Viktor Zaverkin}, \bibinfo{person}{Takashi Maruyama}, \bibinfo{person}{Mathias Niepert}, {and} \bibinfo{person}{Francesco Alesiani}.} \bibinfo{year}{2023}\natexlab{}.
\newblock \showarticletitle{Adaptive Message Passing: A General Framework to Mitigate Oversmoothing, Oversquashing, and Underreaching}.
\newblock \bibinfo{journal}{\emph{arXiv preprint arXiv:2312.16560}} (\bibinfo{year}{2023}).
\newblock


\bibitem[Faber and Wattenhofer(2024)]%
        {faber2024gwac}
\bibfield{author}{\bibinfo{person}{Lukas Faber} {and} \bibinfo{person}{Roger Wattenhofer}.} \bibinfo{year}{2024}\natexlab{}.
\newblock \showarticletitle{GwAC: GNNs with Asynchronous Communication}. In \bibinfo{booktitle}{\emph{Learning on Graphs Conference}}. PMLR, \bibinfo{pages}{8--1}.
\newblock


\bibitem[Finkelshtein et~al\mbox{.}(2024)]%
        {finkelshtein2023cooperative}
\bibfield{author}{\bibinfo{person}{Ben Finkelshtein}, \bibinfo{person}{Xingyue Huang}, \bibinfo{person}{Michael~M. Bronstein}, {and} \bibinfo{person}{Ismail~Ilkan Ceylan}.} \bibinfo{year}{2024}\natexlab{}.
\newblock \bibinfo{title}{Cooperative Graph Neural Networks}.
\newblock
\urldef\tempurl%
\url{https://openreview.net/forum?id=T0FuEDnODP}
\showURL{%
\tempurl}


\bibitem[Gilmer et~al\mbox{.}(2017)]%
        {gilmer2017neural}
\bibfield{author}{\bibinfo{person}{Justin Gilmer}, \bibinfo{person}{Samuel~S Schoenholz}, \bibinfo{person}{Patrick~F Riley}, \bibinfo{person}{Oriol Vinyals}, {and} \bibinfo{person}{George~E Dahl}.} \bibinfo{year}{2017}\natexlab{}.
\newblock \showarticletitle{Neural message passing for quantum chemistry}. In \bibinfo{booktitle}{\emph{International conference on machine learning}}. PMLR, \bibinfo{pages}{1263--1272}.
\newblock


\bibitem[Giraldo et~al\mbox{.}(2023)]%
        {giraldo2023sjlr}
\bibfield{author}{\bibinfo{person}{Jhony~H. Giraldo}, \bibinfo{person}{Konstantinos Skianis}, \bibinfo{person}{Thierry Bouwmans}, {and} \bibinfo{person}{Fragkiskos~D. Malliaros}.} \bibinfo{year}{2023}\natexlab{}.
\newblock \showarticletitle{On the Trade-off between Over-smoothing and Over-squashing in Deep Graph Neural Networks}. In \bibinfo{booktitle}{\emph{Proceedings of the 32nd ACM International Conference on Information and Knowledge Management}} \emph{(\bibinfo{series}{CIKM '23})}. \bibinfo{pages}{566–576}.
\newblock


\bibitem[Graves(2016)]%
        {graves2016adaptive}
\bibfield{author}{\bibinfo{person}{Alex Graves}.} \bibinfo{year}{2016}\natexlab{}.
\newblock \showarticletitle{Adaptive computation time for recurrent neural networks}.
\newblock \bibinfo{journal}{\emph{arXiv preprint arXiv:1603.08983}} (\bibinfo{year}{2016}).
\newblock


\bibitem[Hu et~al\mbox{.}(2020)]%
        {hu2020open}
\bibfield{author}{\bibinfo{person}{Weihua Hu}, \bibinfo{person}{Matthias Fey}, \bibinfo{person}{Marinka Zitnik}, \bibinfo{person}{Yuxiao Dong}, \bibinfo{person}{Hongyu Ren}, \bibinfo{person}{Bowen Liu}, \bibinfo{person}{Michele Catasta}, {and} \bibinfo{person}{Jure Leskovec}.} \bibinfo{year}{2020}\natexlab{}.
\newblock \showarticletitle{Open graph benchmark: Datasets for machine learning on graphs}.
\newblock \bibinfo{journal}{\emph{Advances in neural information processing systems}}  \bibinfo{volume}{33} (\bibinfo{year}{2020}), \bibinfo{pages}{22118--22133}.
\newblock


\bibitem[Huang et~al\mbox{.}(2016)]%
        {huang2016deep}
\bibfield{author}{\bibinfo{person}{Gao Huang}, \bibinfo{person}{Yu Sun}, \bibinfo{person}{Zhuang Liu}, \bibinfo{person}{Daniel Sedra}, {and} \bibinfo{person}{Kilian~Q Weinberger}.} \bibinfo{year}{2016}\natexlab{}.
\newblock \showarticletitle{Deep networks with stochastic depth}. In \bibinfo{booktitle}{\emph{Computer Vision--ECCV 2016: 14th European Conference, Amsterdam, The Netherlands, October 11--14, 2016, Proceedings, Part IV 14}}. Springer, \bibinfo{pages}{646--661}.
\newblock


\bibitem[Hur{\'e} et~al\mbox{.}(2021)]%
        {hure2021deep}
\bibfield{author}{\bibinfo{person}{C{\^o}me Hur{\'e}}, \bibinfo{person}{Huy{\^e}n Pham}, \bibinfo{person}{Achref Bachouch}, {and} \bibinfo{person}{Nicolas Langren{\'e}}.} \bibinfo{year}{2021}\natexlab{}.
\newblock \showarticletitle{Deep neural networks algorithms for stochastic control problems on finite horizon: convergence analysis}.
\newblock \bibinfo{journal}{\emph{SIAM J. Numer. Anal.}} \bibinfo{volume}{59}, \bibinfo{number}{1} (\bibinfo{year}{2021}), \bibinfo{pages}{525--557}.
\newblock


\bibitem[Jia et~al\mbox{.}(2016)]%
        {jia2016dynamic}
\bibfield{author}{\bibinfo{person}{Xu Jia}, \bibinfo{person}{Bert De~Brabandere}, \bibinfo{person}{Tinne Tuytelaars}, {and} \bibinfo{person}{Luc~V Gool}.} \bibinfo{year}{2016}\natexlab{}.
\newblock \showarticletitle{Dynamic filter networks}.
\newblock \bibinfo{journal}{\emph{Advances in neural information processing systems}}  \bibinfo{volume}{29} (\bibinfo{year}{2016}).
\newblock


\bibitem[Kingma and Ba(2014)]%
        {kingma_adam}
\bibfield{author}{\bibinfo{person}{Diederik~P. Kingma} {and} \bibinfo{person}{Jimmy Ba}.} \bibinfo{year}{2014}\natexlab{}.
\newblock \bibinfo{title}{Adam: A Method for Stochastic Optimization}.
\newblock
\href{https://doi.org/10.48550/ARXIV.1412.6980}{doi:\nolinkurl{10.48550/ARXIV.1412.6980}}


\bibitem[Kipf and Welling(2017)]%
        {Kipf:2017tc}
\bibfield{author}{\bibinfo{person}{Thomas~N. Kipf} {and} \bibinfo{person}{Max Welling}.} \bibinfo{year}{2017}\natexlab{}.
\newblock \showarticletitle{{Semi-Supervised Classification with Graph Convolutional Networks}}. In \bibinfo{booktitle}{\emph{ICLR}}.
\newblock


\bibitem[Li et~al\mbox{.}(2017)]%
        {li2017pruning}
\bibfield{author}{\bibinfo{person}{Hao Li}, \bibinfo{person}{Asim Kadav}, \bibinfo{person}{Igor Durdanovic}, \bibinfo{person}{Hanan Samet}, {and} \bibinfo{person}{Hans~Peter Graf}.} \bibinfo{year}{2017}\natexlab{}.
\newblock \showarticletitle{Pruning Filters for Efficient ConvNets}. In \bibinfo{booktitle}{\emph{International Conference on Learning Representations}}.
\newblock
\urldef\tempurl%
\url{https://openreview.net/forum?id=rJqFGTslg}
\showURL{%
\tempurl}


\bibitem[Lim and Benson(2021)]%
        {lim2021expertise}
\bibfield{author}{\bibinfo{person}{Derek Lim} {and} \bibinfo{person}{Austin~R Benson}.} \bibinfo{year}{2021}\natexlab{}.
\newblock \showarticletitle{Expertise and dynamics within crowdsourced musical knowledge curation: A case study of the genius platform}. In \bibinfo{booktitle}{\emph{Proceedings of the International AAAI Conference on Web and Social Media}}, Vol.~\bibinfo{volume}{15}. \bibinfo{pages}{373--384}.
\newblock


\bibitem[Lim et~al\mbox{.}(2021)]%
        {lim2021large}
\bibfield{author}{\bibinfo{person}{Derek Lim}, \bibinfo{person}{Felix~Matthew Hohne}, \bibinfo{person}{Xiuyu Li}, \bibinfo{person}{Sijia~Linda Huang}, \bibinfo{person}{Vaishnavi Gupta}, \bibinfo{person}{Omkar~Prasad Bhalerao}, {and} \bibinfo{person}{Ser-Nam Lim}.} \bibinfo{year}{2021}\natexlab{}.
\newblock \showarticletitle{Large Scale Learning on Non-Homophilous Graphs: New Benchmarks and Strong Simple Methods}. In \bibinfo{booktitle}{\emph{Advances in Neural Information Processing Systems}}, \bibfield{editor}{\bibinfo{person}{A.~Beygelzimer}, \bibinfo{person}{Y.~Dauphin}, \bibinfo{person}{P.~Liang}, {and} \bibinfo{person}{J.~Wortman Vaughan}} (Eds.).
\newblock


\bibitem[Liu et~al\mbox{.}(2021)]%
        {liu2021eignn}
\bibfield{author}{\bibinfo{person}{Juncheng Liu}, \bibinfo{person}{Kenji Kawaguchi}, \bibinfo{person}{Bryan Hooi}, \bibinfo{person}{Yiwei Wang}, {and} \bibinfo{person}{Xiaokui Xiao}.} \bibinfo{year}{2021}\natexlab{}.
\newblock \showarticletitle{Eignn: Efficient infinite-depth graph neural networks}.
\newblock \bibinfo{journal}{\emph{Advances in Neural Information Processing Systems}}  \bibinfo{volume}{34} (\bibinfo{year}{2021}), \bibinfo{pages}{18762--18773}.
\newblock


\bibitem[Luan et~al\mbox{.}(2022)]%
        {luan2022revisiting}
\bibfield{author}{\bibinfo{person}{Sitao Luan}, \bibinfo{person}{Chenqing Hua}, \bibinfo{person}{Qincheng Lu}, \bibinfo{person}{Jiaqi Zhu}, \bibinfo{person}{Mingde Zhao}, \bibinfo{person}{Shuyuan Zhang}, \bibinfo{person}{Xiao-Wen Chang}, {and} \bibinfo{person}{Doina Precup}.} \bibinfo{year}{2022}\natexlab{}.
\newblock \showarticletitle{Revisiting heterophily for graph neural networks}.
\newblock \bibinfo{journal}{\emph{Advances in neural information processing systems}}  \bibinfo{volume}{35} (\bibinfo{year}{2022}), \bibinfo{pages}{1362--1375}.
\newblock


\bibitem[Malliaros et~al\mbox{.}(2020)]%
        {kcore-vldbj20}
\bibfield{author}{\bibinfo{person}{Fragkiskos~D. Malliaros}, \bibinfo{person}{Christos Giatsidis}, \bibinfo{person}{Apostolos~N. Papadopoulos}, {and} \bibinfo{person}{Michalis Vazirgiannis}.} \bibinfo{year}{2020}\natexlab{}.
\newblock \showarticletitle{The core decomposition of networks: theory, algorithms and applications}.
\newblock \bibinfo{journal}{\emph{{VLDB} J.}} \bibinfo{volume}{29}, \bibinfo{number}{1} (\bibinfo{year}{2020}), \bibinfo{pages}{61--92}.
\newblock


\bibitem[Malliaros and Vazirgiannis(2013)]%
        {MALLIAROS201395}
\bibfield{author}{\bibinfo{person}{Fragkiskos~D. Malliaros} {and} \bibinfo{person}{Michalis Vazirgiannis}.} \bibinfo{year}{2013}\natexlab{}.
\newblock \showarticletitle{Clustering and community detection in directed networks: A survey}.
\newblock \bibinfo{journal}{\emph{Physics Reports}} \bibinfo{volume}{533}, \bibinfo{number}{4} (\bibinfo{year}{2013}), \bibinfo{pages}{95--142}.
\newblock
\newblock
\shownote{Clustering and Community Detection in Directed Networks: A Survey}.


\bibitem[Panagopoulos et~al\mbox{.}(2024)]%
        {glie-asonam24}
\bibfield{author}{\bibinfo{person}{George Panagopoulos}, \bibinfo{person}{Nikolaos Tziortziotis}, \bibinfo{person}{Michalis Vazirgiannis}, {and} \bibinfo{person}{Fragkiskos Malliaros}.} \bibinfo{year}{2024}\natexlab{}.
\newblock \showarticletitle{Maximizing Influence with Graph Neural Networks}. In \bibinfo{booktitle}{\emph{Proceedings of the 2023 IEEE/ACM International Conference on Advances in Social Networks Analysis and Mining}}. \bibinfo{pages}{237–244}.
\newblock


\bibitem[Ramp{\'a}{\v{s}}ek et~al\mbox{.}(2022)]%
        {rampavsek2022recipe}
\bibfield{author}{\bibinfo{person}{Ladislav Ramp{\'a}{\v{s}}ek}, \bibinfo{person}{Michael Galkin}, \bibinfo{person}{Vijay~Prakash Dwivedi}, \bibinfo{person}{Anh~Tuan Luu}, \bibinfo{person}{Guy Wolf}, {and} \bibinfo{person}{Dominique Beaini}.} \bibinfo{year}{2022}\natexlab{}.
\newblock \showarticletitle{Recipe for a general, powerful, scalable graph transformer}.
\newblock \bibinfo{journal}{\emph{Advances in Neural Information Processing Systems}}  \bibinfo{volume}{35} (\bibinfo{year}{2022}), \bibinfo{pages}{14501--14515}.
\newblock


\bibitem[Rozemberczki et~al\mbox{.}(2021)]%
        {rozemberczki2021multi}
\bibfield{author}{\bibinfo{person}{Benedek Rozemberczki}, \bibinfo{person}{Carl Allen}, {and} \bibinfo{person}{Rik Sarkar}.} \bibinfo{year}{2021}\natexlab{}.
\newblock \showarticletitle{Multi-scale attributed node embedding}.
\newblock \bibinfo{journal}{\emph{Journal of Complex Networks}} \bibinfo{volume}{9}, \bibinfo{number}{2} (\bibinfo{year}{2021}), \bibinfo{pages}{cnab014}.
\newblock


\bibitem[Sabour et~al\mbox{.}(2017)]%
        {sabour2017dynamic}
\bibfield{author}{\bibinfo{person}{Sara Sabour}, \bibinfo{person}{Nicholas Frosst}, {and} \bibinfo{person}{Geoffrey~E Hinton}.} \bibinfo{year}{2017}\natexlab{}.
\newblock \showarticletitle{Dynamic routing between capsules}.
\newblock \bibinfo{journal}{\emph{Advances in neural information processing systems}}  \bibinfo{volume}{30} (\bibinfo{year}{2017}).
\newblock


\bibitem[Schuster et~al\mbox{.}(2022)]%
        {schuster2022confident}
\bibfield{author}{\bibinfo{person}{Tal Schuster}, \bibinfo{person}{Adam Fisch}, \bibinfo{person}{Jai Gupta}, \bibinfo{person}{Mostafa Dehghani}, \bibinfo{person}{Dara Bahri}, \bibinfo{person}{Vinh~Q. Tran}, \bibinfo{person}{Yi Tay}, {and} \bibinfo{person}{Donald Metzler}.} \bibinfo{year}{2022}\natexlab{}.
\newblock \showarticletitle{Confident Adaptive Language Modeling}. In \bibinfo{booktitle}{\emph{Advances in Neural Information Processing Systems}}, \bibfield{editor}{\bibinfo{person}{Alice~H. Oh}, \bibinfo{person}{Alekh Agarwal}, \bibinfo{person}{Danielle Belgrave}, {and} \bibinfo{person}{Kyunghyun Cho}} (Eds.).
\newblock
\urldef\tempurl%
\url{https://openreview.net/forum?id=uLYc4L3C81A}
\showURL{%
\tempurl}


\bibitem[Sen et~al\mbox{.}(2008)]%
        {dataset_node_classification}
\bibfield{author}{\bibinfo{person}{Prithviraj Sen}, \bibinfo{person}{Galileo Namata}, \bibinfo{person}{Mustafa Bilgic}, \bibinfo{person}{Lise Getoor}, \bibinfo{person}{Brian Galligher}, {and} \bibinfo{person}{Tina Eliassi-Rad}.} \bibinfo{year}{2008}\natexlab{}.
\newblock \showarticletitle{Collective Classification in Network Data}.
\newblock \bibinfo{journal}{\emph{AI Magazine}} \bibinfo{volume}{29}, \bibinfo{number}{3} (\bibinfo{date}{Sep.} \bibinfo{year}{2008}), \bibinfo{pages}{93}.
\newblock
\href{https://doi.org/10.1609/aimag.v29i3.2157}{doi:\nolinkurl{10.1609/aimag.v29i3.2157}}


\bibitem[Shchur et~al\mbox{.}(2018)]%
        {cs_data}
\bibfield{author}{\bibinfo{person}{Oleksandr Shchur}, \bibinfo{person}{Maximilian Mumme}, \bibinfo{person}{Aleksandar Bojchevski}, {and} \bibinfo{person}{Stephan Günnemann}.} \bibinfo{year}{2018}\natexlab{}.
\newblock \showarticletitle{Pitfalls of Graph Neural Network Evaluation}. In \bibinfo{booktitle}{\emph{NeurIPS Relational Representation Learning Workshop (R2L 2018)}}.
\newblock


\bibitem[Spinelli et~al\mbox{.}(2020)]%
        {spinelli2020adaptive}
\bibfield{author}{\bibinfo{person}{Indro Spinelli}, \bibinfo{person}{Simone Scardapane}, {and} \bibinfo{person}{Aurelio Uncini}.} \bibinfo{year}{2020}\natexlab{}.
\newblock \showarticletitle{Adaptive propagation graph convolutional network}.
\newblock \bibinfo{journal}{\emph{IEEE Transactions on Neural Networks and Learning Systems}} \bibinfo{volume}{32}, \bibinfo{number}{10} (\bibinfo{year}{2020}), \bibinfo{pages}{4755--4760}.
\newblock


\bibitem[Sun et~al\mbox{.}(2021)]%
        {sun2019adagcn}
\bibfield{author}{\bibinfo{person}{Ke Sun}, \bibinfo{person}{Zhanxing Zhu}, {and} \bibinfo{person}{Zhouchen Lin}.} \bibinfo{year}{2021}\natexlab{}.
\newblock \showarticletitle{Ada{\{}GCN{\}}: Adaboosting Graph Convolutional Networks into Deep Models}. In \bibinfo{booktitle}{\emph{International Conference on Learning Representations}}.
\newblock
\urldef\tempurl%
\url{https://openreview.net/forum?id=QkRbdiiEjM}
\showURL{%
\tempurl}


\bibitem[Vaswani et~al\mbox{.}(2017)]%
        {DBLP:journals/corr/VaswaniSPUJGKP17}
\bibfield{author}{\bibinfo{person}{Ashish Vaswani}, \bibinfo{person}{Noam Shazeer}, \bibinfo{person}{Niki Parmar}, \bibinfo{person}{Jakob Uszkoreit}, \bibinfo{person}{Llion Jones}, \bibinfo{person}{Aidan~N. Gomez}, \bibinfo{person}{Lukasz Kaiser}, {and} \bibinfo{person}{Illia Polosukhin}.} \bibinfo{year}{2017}\natexlab{}.
\newblock \showarticletitle{Attention Is All You Need}.
\newblock \bibinfo{journal}{\emph{Advances in neural information processing systems}} (\bibinfo{year}{2017}).
\newblock


\bibitem[Wang et~al\mbox{.}(2020)]%
        {DBLP:journals/corr/abs-2010-05300}
\bibfield{author}{\bibinfo{person}{Yulin Wang}, \bibinfo{person}{Kangchen Lv}, \bibinfo{person}{Rui Huang}, \bibinfo{person}{Shiji Song}, \bibinfo{person}{Le Yang}, {and} \bibinfo{person}{Gao Huang}.} \bibinfo{year}{2020}\natexlab{}.
\newblock \showarticletitle{Glance and Focus: a Dynamic Approach to Reducing Spatial Redundancy in Image Classification}.
\newblock \bibinfo{journal}{\emph{CoRR}}  \bibinfo{volume}{abs/2010.05300} (\bibinfo{year}{2020}).
\newblock
\showeprint[arXiv]{2010.05300}
\urldef\tempurl%
\url{https://arxiv.org/abs/2010.05300}
\showURL{%
\tempurl}


\bibitem[Xu et~al\mbox{.}(2019)]%
        {xu2019powerful}
\bibfield{author}{\bibinfo{person}{Keyulu Xu}, \bibinfo{person}{Weihua Hu}, \bibinfo{person}{Jure Leskovec}, {and} \bibinfo{person}{Stefanie Jegelka}.} \bibinfo{year}{2019}\natexlab{}.
\newblock \showarticletitle{{How Powerful are Graph Neural Networks?}}. In \bibinfo{booktitle}{\emph{7th International Conference on Learning Representations}}.
\newblock


\bibitem[Xu et~al\mbox{.}(2018a)]%
        {DBLP:journals/corr/abs-1806-03536}
\bibfield{author}{\bibinfo{person}{Keyulu Xu}, \bibinfo{person}{Chengtao Li}, \bibinfo{person}{Yonglong Tian}, \bibinfo{person}{Tomohiro Sonobe}, \bibinfo{person}{Ken{-}ichi Kawarabayashi}, {and} \bibinfo{person}{Stefanie Jegelka}.} \bibinfo{year}{2018}\natexlab{a}.
\newblock \showarticletitle{Representation Learning on Graphs with Jumping Knowledge Networks}.
\newblock \bibinfo{journal}{\emph{CoRR}}  \bibinfo{volume}{abs/1806.03536} (\bibinfo{year}{2018}).
\newblock
\showeprint[arXiv]{1806.03536}
\urldef\tempurl%
\url{http://arxiv.org/abs/1806.03536}
\showURL{%
\tempurl}


\bibitem[Xu et~al\mbox{.}(2018b)]%
        {pmlr-v80-xu18c}
\bibfield{author}{\bibinfo{person}{Keyulu Xu}, \bibinfo{person}{Chengtao Li}, \bibinfo{person}{Yonglong Tian}, \bibinfo{person}{Tomohiro Sonobe}, \bibinfo{person}{Ken-ichi Kawarabayashi}, {and} \bibinfo{person}{Stefanie Jegelka}.} \bibinfo{year}{2018}\natexlab{b}.
\newblock \showarticletitle{Representation learning on graphs with jumping knowledge networks}. In \bibinfo{booktitle}{\emph{International conference on machine learning}}. pmlr, \bibinfo{pages}{5453--5462}.
\newblock


\bibitem[Zeng et~al\mbox{.}(2020)]%
        {zeng2020deep}
\bibfield{author}{\bibinfo{person}{Hanqing Zeng}, \bibinfo{person}{Muhan Zhang}, \bibinfo{person}{Yinglong Xia}, \bibinfo{person}{Ajitesh Srivastava}, \bibinfo{person}{Rajgopal Kannan}, \bibinfo{person}{Viktor Prasanna}, \bibinfo{person}{Long Jin}, \bibinfo{person}{Andrey Malevich}, {and} \bibinfo{person}{Ren Chen}.} \bibinfo{year}{2020}\natexlab{}.
\newblock \showarticletitle{Deep graph neural networks with shallow subgraph samplers}.
\newblock  (\bibinfo{year}{2020}).
\newblock


\bibitem[Zhang et~al\mbox{.}(2021)]%
        {zhang2021evaluating}
\bibfield{author}{\bibinfo{person}{Wentao Zhang}, \bibinfo{person}{Zeang Sheng}, \bibinfo{person}{Yuezihan Jiang}, \bibinfo{person}{Yikuan Xia}, \bibinfo{person}{Jun Gao}, \bibinfo{person}{Zhi Yang}, {and} \bibinfo{person}{Bin Cui}.} \bibinfo{year}{2021}\natexlab{}.
\newblock \showarticletitle{Evaluating deep graph neural networks}.
\newblock \bibinfo{journal}{\emph{arXiv preprint arXiv:2108.00955}} (\bibinfo{year}{2021}).
\newblock


\bibitem[Zhao and Akoglu(2020)]%
        {Zhao2020PairNorm}
\bibfield{author}{\bibinfo{person}{Lingxiao Zhao} {and} \bibinfo{person}{Leman Akoglu}.} \bibinfo{year}{2020}\natexlab{}.
\newblock \showarticletitle{PairNorm: Tackling Oversmoothing in GNNs}. In \bibinfo{booktitle}{\emph{International Conference on Learning Representations}}.
\newblock


\bibitem[Zhou et~al\mbox{.}(2020)]%
        {DBLP:journals/corr/abs-2006-04152}
\bibfield{author}{\bibinfo{person}{Wangchunshu Zhou}, \bibinfo{person}{Canwen Xu}, \bibinfo{person}{Tao Ge}, \bibinfo{person}{Julian~J. McAuley}, \bibinfo{person}{Ke Xu}, {and} \bibinfo{person}{Furu Wei}.} \bibinfo{year}{2020}\natexlab{}.
\newblock \showarticletitle{{BERT} Loses Patience: Fast and Robust Inference with Early Exit}.
\newblock \bibinfo{journal}{\emph{CoRR}}  \bibinfo{volume}{abs/2006.04152} (\bibinfo{year}{2020}).
\newblock
\showeprint[arXiv]{2006.04152}
\urldef\tempurl%
\url{https://arxiv.org/abs/2006.04152}
\showURL{%
\tempurl}


\end{thebibliography}
\end{document}